\documentclass[preprint,12pt]{elsarticle}
\graphicspath{{./figures/}}

\usepackage{fontspec}
\usepackage[bidi=default,english]{babel}
\usepackage{needspace}

\babelfont{rm}[
  UprightFont = lmroman12-regular.otf,
  BoldFont = lmroman12-bold.otf,
  ItalicFont = lmroman12-italic.otf
]{LatinModernRoman}
\babelprovide[import,main]{english}
\babelprovide[import]{arabic}
\babelfont[*arabic]{rm}[
  UprightFont = Amiri-Regular.ttf,
  BoldFont = Amiri-Bold.ttf,
  ItalicFont = Amiri-Italic.ttf,
  BoldItalicFont = Amiri-BoldItalic.ttf
]{Amiri}

\providecommand{\textLR}[1]{\foreignlanguage{english}{#1}}

\usepackage[table]{xcolor}

\definecolor{jsonbg}{HTML}{F6F8FA}
\definecolor{jsonframe}{HTML}{D0D7DE}
\definecolor{boxgray}{HTML}{F3F4F6}

\usepackage{booktabs}
\usepackage{array}
\usepackage{ragged2e}
\usepackage{siunitx}

\newcommand{\arcell}[1]{\hfill\foreignlanguage{arabic}{#1}}

\usepackage[most]{tcolorbox}
\tcbuselibrary{skins,breakable,listings}
\usepackage{listings}

\tcbset{
  myarbox/.style={
    enhanced,
    breakable,
    colback=boxgray,
    colframe=gray!35,
    boxrule=0.35pt,
    arc=1.2mm,
    left=5pt,
    right=5pt,
    top=5pt,
    bottom=5pt,
    before skip=6pt,
    after skip=6pt
  }
}

\newtcblisting{jsonbox}{
  enhanced,
  breakable,
  listing only,
  colback=jsonbg,
  colframe=jsonframe,
  boxrule=0.4pt,
  arc=2mm,
  left=6pt,
  right=6pt,
  top=6pt,
  bottom=6pt,
  before skip=8pt,
  after skip=8pt,
  listing options={
    basicstyle=\ttfamily\footnotesize,
    breaklines=true,
    columns=fullflexible,
    keepspaces=true,
    showstringspaces=false,
    frame=none
  }
}

\usepackage{enumitem}
\usepackage{amssymb}
\usepackage{amsmath}
\usepackage{float}
\usepackage{verbatim}
\usepackage{xurl}
\usepackage{hyperref}

\hypersetup{
  colorlinks=true,
  urlcolor=blue,
  linkcolor=black,
  citecolor=black
}

\restylefloat{figure}
\floatstyle{plaintop}
\restylefloat{table}

\journal{Expert Systems with Applications}

\begin{document}
\author[hbku]{Abdessalam BOUCHEKIF}
\author[hbku]{Shahd GABEN}
\author[hbku]{Samer RASHWANI}
\author[hbku]{Somaya ELTANBOULY}
\author[hbku]{Mutaz AL-KHATIB}
\author[hbku]{Heba SBAHI}
\author[hbku]{Mohammed GHALY}
\author[nu]{Emad MOHAMED}

\address[hbku]{Hamad Bin Khalifa University, Qatar}
\address[nu]{Nazarbayev University, Kazakhstan}
\begin{frontmatter}

\title{MAWARITH: A Dataset and Benchmark for Legal Inheritance
Reasoning with LLMs}

\begin{abstract}
Islamic inheritance law (\textit{ʿilm al-mawārīth}) is challenging for large language models because solving inheritance cases requires complex, structured, multi-step reasoning and the correct application of juristic rules to compute heirs’ shares. We introduce \textit{MAWARITH}, a large-scale annotated dataset of $12{,}500$ Arabic inheritance cases for training and evaluating models on the full reasoning chain: \textit{(i)} identifying eligible heirs, \textit{(ii)} applying blocking (\textit{ḥajb}) and allocation rules, and \textit{(iii)} computing exact inheritance shares. To the best of our knowledge, \textit{MAWARITH} is the first Arabic corpus and benchmark designed for end-to-end Islamic inheritance reasoning. Unlike prior datasets that restrict inheritance case solving to multiple-choice questions, \textit{MAWARITH} supports the full reasoning chain and provides step-by-step solutions with justifications grounded in classical juristic sources and established inheritance rules, as well as exact share calculations. This enables models to learn how to generate detailed, step-by-step responses to user queries that reflect real-world Islamic inheritance cases. To evaluate models beyond final-answer accuracy, we propose \textit{MIR-E} (Mawarith Inheritance Reasoning Evaluation), a weighted multi-stage metric that scores key reasoning stages and captures error propagation across the pipeline. We evaluate six large language models in a zero-shot setting. A commercial model achieves about $90\%$, whereas all evaluated open-source models remain below $50\%$. Our error analysis identifies recurring failure patterns, including scenario misinterpretation, errors in heir identification, errors in share allocation, and missing or incorrect application of key inheritance rules such as \textit{ʿawl} and \textit{radd}. The \textit{MAWARITH} dataset is publicly available at \textLR{\url{https://gitlab.com/nlpresearcher/mawarith}}.

\end{abstract}


\end{frontmatter}

\section{Introduction}
Large language models (LLMs) have recently achieved strong performance across a wide range of natural language processing tasks, 
including question answering, summarization, and legal text understanding. 
Despite these advances, their ability to perform \emph{structured, rule-based reasoning} 
remains limited and is still insufficiently evaluated. 
For some tasks, fluent generation and general knowledge are not sufficient;  models must follow precise reasoning steps, where early errors  can propagate and invalidate subsequent decisions.
This limitation has motivated the development of models designed to solve complex tasks by breaking them into smaller steps and producing intermediate reasoning traces before generating a final output.
Recent efforts in this direction include reasoning-oriented large language models such as the latest versions of Gemini, GPT, DeepSeek, Fanar, and Qwen,
which aim to improve multi-step reasoning by decomposing the problem into smaller steps, generating intermediate steps, verifying intermediate results, and correcting errors when needed.
The evaluation of these models has largely focused on mathematical and synthetic reasoning benchmarks (e.g., GSM8K, MATH, and BBH; [1, 2, 3]),
while structured domain reasoning remains less covered, despite dedicated efforts in law
[4, 5, 6]
and medicine [7, 8].
Islamic inheritance law (\textit{ʿilm al-mawārīth}) offers a particularly challenging and well-defined testbed 
for evaluating such reasoning capabilities. 
Solving inheritance problems in this domain requires multi-step legal reasoning. A correct solution must first identify the eligible heirs, 
then apply the juristic rules governing exclusion and blocking, 
and finally compute the exact fractional shares. The process is further complicated by adjustment cases such as \textit{ʿawl} and \textit{radd}.

In this work, we introduce a large-scale benchmark corpus of 12,500 Islamic inheritance problems, designed to support training  and evaluation  models for multi-step legal–numerical reasoning under strict jurisprudential rules. Each case is annotated with detailed, human-like reasoning that mirrors the step-by-step process followed by jurists: including heir identification, blocking analysis, and systematic share computation.
The corpus covers a wide range of inheritance scenarios with varying levels of complexity. To enable fine-grained evaluation, we adopt a multi-stage evaluation framework that assesses not only the correctness of the final numerical shares, but also the validity of the intermediate reasoning steps involved in reaching the solution. This evaluation strategy allows for a precise analysis of reasoning errors and failure modes across the core stages of inheritance problem solving.
\\
The remainder of this paper is organized as follows. Section \ref{rl} reviews the related work. Section \ref{background} provides an overview of Islamic inheritance law. Section \ref{data} describes the dataset used in this study. Section \ref{eval} details the evaluation metrics, and Section \ref{exp} outlines the experimental setup. Section \ref{analysis} presents an in-depth error analysis. Finally, Section \ref{sec:conclusion} concludes the paper and discusses potential directions for future research.

\section{Related Work} \label{rl}

Large language models  have recently been applied to a wide range of Islamic knowledge tasks, 
including Quranic question answering~[9], 
knowledge retrieval~[10, 11], 
and the analysis of hallucinations in Islamic content~[10]. 
These studies show that LLMs perform well on knowledge retrieval and basic understanding when answers rely on direct textual matching. However, they often hallucinate and show clear limitations on tasks that require structured reasoning or deep domain knowledge.
In [12], the authors report that several models, such as LLaMA and ALLaM, frequently cite non-existent Quranic verses or fabricate Hadith references.
As a result, the generated conclusions are not only incorrect, 
but are also supported by false religious evidence. 
This behavior raises serious concerns for religious and legal applications, 
where correctness depends not only on the final answer 
but also on the authenticity and reliability of the cited sources.
\\
To mitigate hallucinations, recent work has explored Retrieval-Augmented Generation (RAG) approaches. 
While RAG improves access to relevant information and enhances factual faithfulness 
and citation accuracy~[9, 13, 14], 
it remains insufficient for answering questions that require multi-step reasoning. 
This limitation has motivated the development of reasoning-oriented models 
that explicitly support multi-step inference. 
In this context, recent research has increasingly focused on such models
that aim to move beyond surface-level text generation 
and  promote more reliable reasoning. 
Models such \textit{o3}, \textit{GPT-5} [15], \textit{Gemini-2.5}~[16], \textit{Gemini3}, \textit{DeepSeek-R1}~[17],  along with open models such as \textit{Fanar-C-2-27B}, \textit{Falcon-H1R}~[18],  \textit{Fanar-Sadiq} [19]
and \textit{Qwen3} ~[20] illustrate this trend 
by promoting more consistent multi-step inference through instruction tuning 
and large-scale pretraining. 
Evaluations of reasoning-oriented language models have largely focused on mathematical and logical benchmarks, on which these models have achieved strong results, particularly in arithmetic reasoning, symbolic manipulation, and competition-style mathematics~[21, 22, 23]. 
Beyond mathematical and logical benchmarks, recent work has begun to investigate LLM reasoning in legally grounded settings by evaluating models on legal benchmarks such as BRIEFME~[24], which require structured argumentation and rule-based reasoning.
The authors show that GPT-4o can outperform human annotators on argument summarization by producing clear and coherent summaries.
Even within the Islamic domain, inheritance law has received growing attention as a challenging testbed for LLM reasoning [25, 26, 27].
In particular, QIAS 2025\footnote{https://sites.google.com/view/qias2025/} [28] a shared task dedicated to Islamic inheritance law (\textit{ʿilm al-mawārīth}), focusing on the evaluation of large language models under strict, rule-based legal and numerical constraints, using a benchmark of $2200$ MCQs.  A similar MCQs benchmark is \textit{MirathQA} [29], built from $1394$ inheritance cases.  Studies report that commercially deployed (\textit{e.g., Gemini and ChatGPT}), reasoning-oriented models consistently outperform non-reasoning or general-purpose models on benchmarks requiring multi-step inference and structured reasoning ~[12, 30, 31, 32, 33]. Additionally,  Elrefai et al.~[34] show that a fine-tuned Qwen3 model achieved top-ranked performance on the   QIAS 2025 shared task. However, this evaluation setup does not allow assessing whether models truly reason correctly. Models were required to select a single correct answer among six options, without any evaluation of the validity of their intermediate reasoning steps or the correctness of the legal justifications leading to that choice. Moreover, Bouchekif et al.~[12] show that even when a model selects the correct answer, the underlying reasoning can still be incorrect or legally invalid. In contrast, the present study requires models to perform end-to-end inheritance reasoning, explicitly generating intermediate reasoning steps, applying jurisprudential rules, and computing the final inheritance shares.

\section{Background: Islamic Inheritance Law} \label{background}
Islamic inheritance  determines the share of each heir according to legal rules derived from the Qur'an, the Prophetic tradition, and juristic interpretation.
Resolving an  inheritance case  requires a well-defined, multi-step procedure that combines juristic rules with precise fractional calculations.
It begins with identifying all potential heirs mentioned in the estate description. 
Eligibility is then assessed: eligible heirs receive their prescribed shares, while others are excluded according to the rules of blocking (ḥajb). 
Two adjustment mechanisms may apply thereafter:
\begin{itemize}
    \item \textit{Radd} (redistribution) applies when the total of the prescribed shares assigned to the heirs is \emph{less than the whole estate} and no residuary heir is present. In this situation, after all fixed shares have been distributed, a portion of the estate remains unassigned. This remaining portion is then redistributed proportionally among the  eligible fixed-share heirs, according to their original shares. 
    For example, if a mother receives $1/6$ of the estate and a daughter receives $1/2$, the total assigned share is $1/6 + 1/2 = 2/3$. Since $2/3$ is less than the full estate, the remaining $1/3$ is redistributed among the mother and the daughter through \textit{radd}.

    \item \textit{ʿAwl} (proportional reduction) applies when the total of the prescribed shares \emph{exceeds the whole estate}. In this case, the initially assigned shares cannot all be satisfied in full. To resolve this, all prescribed shares are proportionally reduced so that their sum becomes exactly equal to the estate. 
    For instance, if the prescribed shares add up to $1/2 + 1/6 + 2/3 = 8/6$, which is greater than $1$, each share is scaled down proportionally through \textit{ʿawl} so that the total distribution fits within the estate.
\end{itemize}
A correct solution requires accurate identification of heirs, strict adherence to juristic rules, and precise numerical computation. Errors at any intermediate stage propagate deterministically and invalidate the final distribution, making \textit{mawārīth} a suitable domain for evaluating structured, multi-step, and rule-based reasoning.

\section{Data Description} \label{data}

We construct a dataset of $12{,}500$ Islamic inheritance problems written in Arabic, following the majority opinion (\textit{al-jumhūr}). Each problem is written as a natural-language inheritance scenario that describes the deceased and the surviving relatives. The corpus covers the full set of kinship relations defined in classical Islamic inheritance law, including direct ascendants and descendants, siblings, spouses, and extended agnatic relatives across multiple generations (e.g., nephews, cousins, uncles, and grandparents). Additional details on the dataset composition, including the distribution of kinship relations, are provided in ~\ref{appendix/appendix_heirs}.

\begin{figure*}[t]
  \centering
  \includegraphics[width=\textwidth]{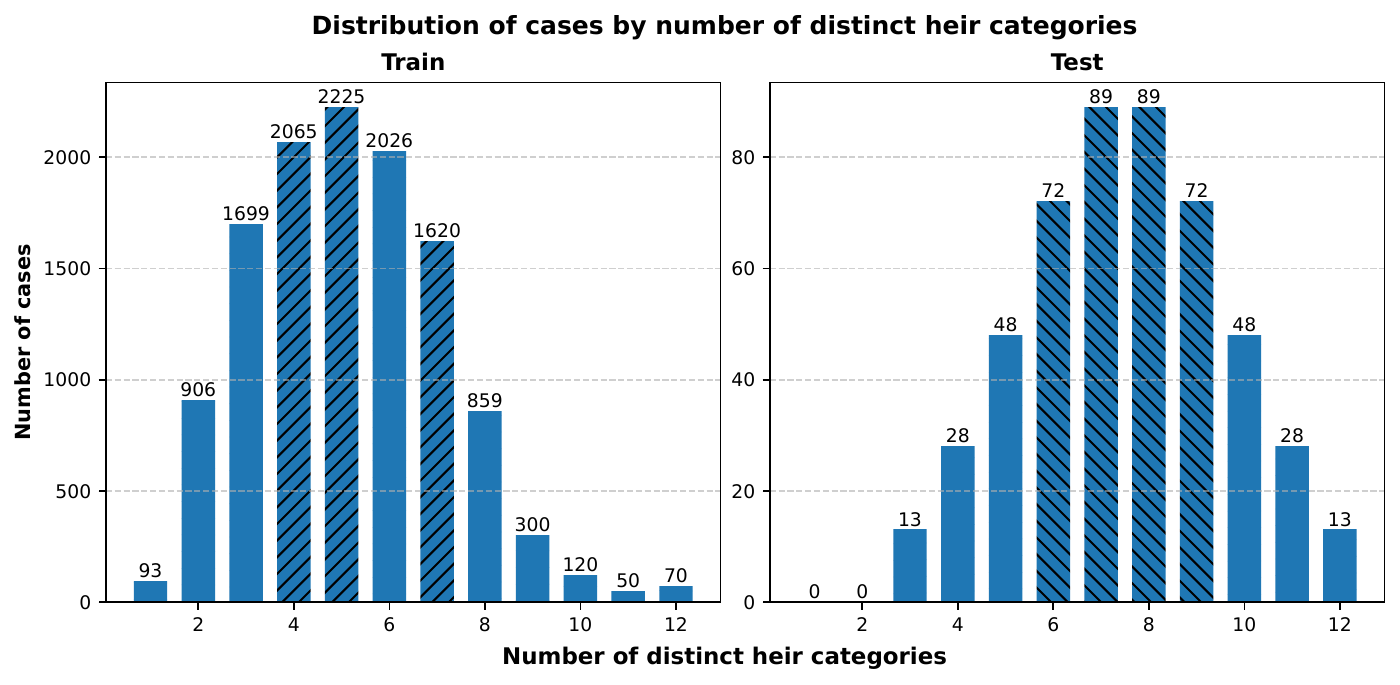}
  \caption{
  Distribution of cases by number of distinct heir categories in the training and test splits.
  Heir categories represent distinct kinship types (e.g., father, mother, siblings). For dataset analysis, multiple individuals within the same kinship group are grouped into a single category.
  The training and test sets share the same distributional profile, differing only in scale.
  }
  \label{fig:heir-categories-distribution}
\end{figure*}

\subsection{Data Collection}
The corpus was constructed through a multi-stage pipeline:
\begin{itemize}
\item \textbf{Case generation.} We first generated inheritance cases using the \textit{Almawarith} inheritance calculator\footnote{\url{https://almwareeth.com/new_masaala}}, which relies on structured, option-based inputs. Eligible heirs were selected via predefined options specifying kinship types and counts. For each structured configuration, the calculator computed the corresponding inheritance shares and sometimes produced brief explanations accompanying the numerical allocations. We use this tool as a reliable source of ground-truth annotations for \textsc{MAWARITH}. While such a calculator can solve the problem once the case has been encoded in a predefined structured format, \textsc{MAWARITH} is an explanation-rich dataset designed to support both training and evaluation on natural-language inheritance cases, including the generation of intermediate reasoning steps and step-by-step explanations.
The output of this stage includes the question, which lists the heirs followed by their counts. For example, the representation \foreignlanguage{arabic}{مات وترك: زوجة (2) وبنت (5) وأخ لأم (2)} indicates that the deceased left behind two wives, five daughters, and two maternal brothers. The output also includes the corresponding share table, taʾṣīl table, and final share table, where each table specifies the heir category, the number of heirs, and the corresponding share.

\item \textbf{Natural-language conversion.} We then converted the structured question into a fluent Arabic natural-language description of the inheritance case, in a form more representative of real-world user queries. For the example above, this conversion produced the following natural-language question:
\begin{quote}
\begin{tcolorbox}[myarbox]
\foreignlanguage{arabic}{
مات وترك: أخوين شقيقين وخمس أخوات شقيقات وأب الأب وأربع بنات ابن ابن.
ما هو نصيب كل وارث؟
}\\
\textit{A person died leaving: two wives, five daughters, and two maternal brothers. What is the share of each heir?}
\end{tcolorbox}
\end{quote}

\item \textbf{Expert enrichment.} The calculator outputs were reviewed and enriched by a two experts in Islamic studies, who added detailed calculations and juristic justifications reflecting the step-by-step reasoning process that a qualified human expert would follow when solving the case.

\item \textbf{Standardization.} The expert explanations were rewritten and standardized using Gemini-2.5-Pro to improve clarity, coherence, and linguistic consistency across the dataset by reducing variations in style, while preserving the underlying juristic reasoning.

\item \textbf{Validation.} Finally, all instances underwent validation checks, including verification of heir counts, consistency between the intermediate reasoning steps and the final shares, and correctness of the numerical computations. Cases involving complex blocking relations or adjustment mechanisms were manually reviewed by two experts, with each case reviewed by a single expert. This review served as a quality-control step to ensure that the Gemini-based reformulation faithfully preserved the original legal reasoning and introduced no substantive errors.
\end{itemize}

\subsection{Reasoning Representation}
\label{struc}
Each dataset instance contains two key components: \texttt{answer} and \textit{answer\_structured}. The \texttt{answer} contains two parts: a detailed intermediate reasoning trace and a final response. The detailed reasoning is enclosed in \textit{<think>} tags, while the final answer is provided within \textit{<answer>} tags.\\
The \texttt{<think>} section contains a step-by-step procedure that describes the reasoning process followed by experts in Islamic inheritance computation. It includes: \textit{(i)} the extraction of potential heirs from the textual description, \textit{(ii)} the determination of eligibility through blocking rules (ḥajb), \textit{(iii)} the assignment of prescribed shares, \textit{(iv)} the application of adjustment mechanisms such as \textit{radd} or \textit{ʿawl} when applicable, and \textit{(v)} the computation of the final inheritance distribution.
\\
\par
\Needspace{18\baselineskip}
\begin{otherlanguage*}{english}
\begin{jsonbox}
{
  "id": "example_id",
  "question": "A person died leaving: ...",
  "answer": "<think> ... </think>\n<answer> ... </answer>",
  "answer_structured": {
    "heirs": [],
    "blocked": [],
    "shares": [],
    "awl_or_radd": "none | awl | radd",
    "awl_stage": {},
    "post_tasil": {}
  }
}
\end{jsonbox}
\end{otherlanguage*}
\par
An illustrative example is provided in ~\ref{app:arabic_example}, showing how a case is solved in \texttt{<think>} and summarized in \texttt{<answer>}. During the heir-identification stage, the \texttt{<think>} section justifies the inclusion (or exclusion) of each candidate heir by applying blocking rules (ḥajb) and may cite relevant Qur'anic evidence when invoking a prescribed-share rule. Then, for each eligible heir, it explains why a specific fraction is assigned and how this fraction depends on the presence or absence of other heirs. The reasoning also makes the numerical procedure explicit: it derives the base denominator (\textit{taʾṣīl}), distributes the residue to the \textit{ʿaṣaba} (residuary heirs), applies correction when needed (\textit{taṣḥīḥ}), and reports final shares as normalized fractions and percentages. Finally, it performs a consistency check by summing all allocated shares to verify that the total equals the full estate (\textit{i.e.}, the shares add up to $1$). The \texttt{<answer>} section then presents the final result in a brief, practical summary format: it reports only the essential final shares assigned to each heir, without repeating the intermediate legal reasoning and numerical steps given in \texttt{<think>}.
For model evaluation, we use a structured format that exposes the outputs of the main reasoning stages: \textit{(i)} the list of heirs who actually inherit, \textit{(ii)} the list of present but blocked heirs, \textit{(iii)} the initial shares before any adjustment, \textit{(iv)} the type of global adjustment applied (\textit{ʿawl} or \textit{radd}), and \textit{(v)} the final normalized distribution (taṣīl). ~\ref{app:answer_structured_format} details the structured output schema of our data and provides a case-general template.

\subsection{Dataset Overview}

The dataset is divided into 12,000 training instances and 500 test instances. It spans a wide range of family configurations, from simple cases with a single heir category to complex scenarios involving up to twelve heir categories.

\begin{table}[!ht]
\centering
\caption{Distribution of inheritance cases by legal complexity}
\label{tab:case_types}

\renewcommand{\arraystretch}{1.15}
\setlength{\tabcolsep}{8pt}

\small
\begin{tabular}{l S S S S}
\toprule
\textbf{Split} & {\textbf{Simple}} & {\textbf{\textit{ʿAwl}}} & {\textbf{Radd}} & {\textbf{Total}} \\
\midrule
Training & 11079 & 577 & 344 & 12000 \\
Test     & 456   & 39  & 5   & 500 \\
\midrule
\textbf{Total} & \bfseries 11535 & \bfseries 616 & \bfseries 349 & \bfseries 12500 \\
\bottomrule
\end{tabular}
\end{table}
The corpus covers 36 distinct heir categories, ranging from close first-degree relatives (such as parents, children, and spouses) to more distant extended family members across multiple generations. These configurations require the application of a wide range of juristic inheritance rules, including fixed shares, residuary principles, blocking relations, and exceptional adjustment cases.
\\
Figure~\ref{fig:heir-categories-distribution} illustrates the distribution of inheritance cases according to the number of distinct heir categories in both the training and test splits. Here, the term \emph{number of heir categories} refers to the number of distinct heir types involved in an inheritance instance, rather than the total number of individual heirs. Given that the corpus is primarily intended for fine-tuning reasoning models, the majority of instances are allocated to the training split in order to maximize exposure to diverse reasoning patterns. In particular, simpler cases involving one or two heir categories are predominantly included in the training set, allowing models to first learn the fundamental inheritance rules before being evaluated on more complex configurations.
\section{Evaluation Metric: MIR-E} \label{eval}
Prior work evaluates reasoning models using either final-answer scoring or LLM-as-a-judge evaluation. However, both strategies have limitations for multi-step reasoning tasks. LLM-based judging can be sensitive to prompt design and difficult to reproduce consistently. Final-answer scoring, in contrast, reduces the entire reasoning process to a set of final values and therefore cannot distinguish between qualitatively different failure modes---for example, a model that makes an early error (e.g., in heir eligibility or blocking) and a model that performs the legal reasoning correctly but makes a numerical error only at the final calculation stage. To address these limitations, we introduce \textbf{MIR-E} (Mawarith Inheritance Reasoning Evaluation), a fine-grained metric designed to assess not only the correctness of the final numerical allocation, but also the validity of intermediate legal reasoning steps. For each inheritance case, models are required to produce outputs in a predefined structured format (as shown in ~\ref{app:answer_structured_format}), enabling detailed comparison with a gold reference solution. Concretely, this format provides structured intermediate outputs for the main reasoning stages, including: \textit{(i)} the list of effective (inheriting) heirs, \textit{(ii)} the list of present but blocked heirs, \textit{(iii)} the initial prescribed shares before any adjustment, \textit{(iv)} the type of global adjustment (\textit{ʿawl} or \textit{radd}), \textit{(v)} the post-adjustment state, and \textit{(vi)} the final normalized distribution (\textit{taʾṣīl}).
\\
MIR-E decomposes inheritance reasoning into four components, each
capturing a distinct source of potential error.

\paragraph{Notation}
Let $H^{gold}$ and $H^{pred}$ denote the sets of heirs identified in the gold
solution and the model prediction, respectively.
Let $B^{gold}$ and $B^{pred}$ denote the sets of heirs that should be blocked
(\textit{ḥajb}) according to the gold solution and the model prediction.
We define the sets of \emph{effective heirs $H_{\text{eff}}$} after blocking as:
\begin{equation*}
\textstyle
H^{gold}_{\text{eff}} = H^{gold} \setminus B^{gold},
\quad
H^{pred}_{\text{eff}} = H^{pred} \setminus B^{pred}.
\end{equation*}

We further define:
\begin{equation*}
\textstyle
H^{miss} = H^{gold}_{\text{eff}} \setminus H^{pred}_{\text{eff}}
\quad \text{(missing heirs)}.
\end{equation*}
\begin{equation*}
\textstyle
H^{spur} = H^{pred}_{\text{eff}} \setminus H^{gold}_{\text{eff}}
\quad \text{(spurious heirs)}.
\end{equation*}
For each heir $h$, let $c_h$ and $\hat{c}_h$ denote the gold and predicted
multiplicities (counts), and let $p_h$ and $\hat{p}_h$ denote the gold and
predicted numerical shares or percentages.
Let $\alpha \in \{\textit{none}, \textit{ʿawl}, \textit{radd}\}$ denote the
normalization regime applied after \textit{taʾṣīl}.

\paragraph{Heirs and Blocking Identification}

This component evaluates whether the model correctly identifies eligible heirs,
blocked heirs, and the count of heirs. We first measure set-level agreement over effective heirs using an F$_1$ score:
\[
{\textstyle
\text{F}_{1}^{\text{heirs}} =
\frac{2 \cdot \lvert H^{gold}_{\text{eff}} \cap H^{pred}_{\text{eff}}\rvert}
{\lvert H^{gold}_{\text{eff}}\rvert + \lvert H^{pred}_{\text{eff}}\rvert}.
}
\]
Explicit penalties are applied for:
\textit{(i)} missing heirs in $H^{miss}$,
\textit{(ii)} spurious heirs in $H^{spur}$,
\textit{(iii)} heirs incorrectly predicted as eligible despite belonging to $B^{gold}$,
and
\textit{(iv)} heirs incorrectly blocked despite being in $H^{gold}_{eff}$

For heirs correctly identified in both sets, we additionally evaluate count
accuracy:
\[
{\textstyle
\text{Acc}_{\text{count}} =
\frac{1}{\lvert H^{gold}_{\text{eff}} \cap H^{pred}_{\text{eff}}\rvert}
\sum_{h \in H^{gold}_{\text{eff}} \cap H^{pred}_{\text{eff}}}
\mathbb{I}(c_h = \hat{c}_h)
}
\]
The final score of this component combines heir identification, blocking
consistency, and multiplicity correctness:
\[
S_{h} = \text{F}_1^{heirs} \cdot \text{Acc}_{count}
\]

\paragraph{Share Assignment}
This component evaluates the correctness of numerical share values assigned to
eligible heirs.
All predicted shares are normalized into numeric values in $[0,1]$.
\\
For each heir $h \in H^{gold}_{eff}$, a predicted share is considered correct if:
\[
|\hat{p}_h - p_h| \leq \varepsilon
\]
where $\varepsilon$ is  tolerance threshold fixed at $0.1$.
\\
The share assignment score is computed as:
\[
{\textstyle
S_{\text{s}} =
\frac{1}{\lvert H^{gold}_{\text{eff}}\rvert}
\sum_{h \in H^{gold}_{\text{eff}}}
\mathbb{I}\!\left(\lvert \hat{p}_h - p_h\rvert \leq \varepsilon\right)
}
\]

\paragraph{Adjustment}
This stage assesses whether normalization is required and whether the model
correctly predicts its type.
Let $\alpha^{gold}, \alpha^{pred} \in \{\text{none}, \text{ʿawl}, \text{radd}\}$
denote the gold and predicted normalization labels.
\\
In our benchmark, 92.3\% of inheritance cases do not require any
normalization adjustment (i.e., $\alpha^{gold} = \text{none}$).
Although normalization is relatively infrequent,
its correct identification remains important,
as both ʿawl (proportional reduction) and radd (redistribution of surplus)
depend directly on the correctness of earlier reasoning steps.
\\
Islamic inheritance reasoning is sequential.
The need for normalization can only be determined
after correctly identifying the effective heirs
and computing their base shares.
If these previous stages are incorrect,
the conditions that justify applying ʿawl or radd are no longer valid.
\\
For this reason, normalization is evaluated conditionally.
The score is defined as:

\[
{\textstyle
S_{\text{a}} =
\begin{cases}
\mathbb{I}(\alpha^{gold} = \alpha^{pred}) &
\text{if } S_{\text{h}} = 1 \text{ and } S_{\text{s}} = 1, \\
0 & \text{otherwise}.
\end{cases}
}
\]

\paragraph{Final Distribution}
This step evaluates whether the model produces the correct final
allocation for each effective heir after the full inheritance process.
For each heir $h \in H^{gold}_{\text{eff}}$, let $f_h$ and $\hat{f}_h$ denote
the gold and predicted final shares. The score is defined as:
\[
{\textstyle
S_{\text{f}} =
\frac{1}{\lvert H^{gold}_{\text{eff}}\rvert}
\sum_{h \in H^{gold}_{\text{eff}}}
\mathbb{I}\!\left(\lvert \hat{f}_h - f_h\rvert \leq \varepsilon\right).
}
\]
\\
where $\varepsilon$ is the tolerance threshold for final-allocation evaluation. In our experiments, we set $\varepsilon = 0.1$.

\paragraph{Overall MIR-E Score}
The final MIR-E score is computed as a weighted sum of the four evaluation
components:
\begin{equation*}
\label{eq:mire}
\text{MIR-E}=\alpha_{\text{h}}\,S_{\text{h}}+\alpha_{\text{s}}\,S_{\text{s}}+\alpha_{\text{a}}\,S_{\text{a}}+\alpha_{\text{f}}\,S_{\text{f}}.
\end{equation*}
We set
$\alpha_{\text{h}} = \alpha_{\text{s}} = \alpha_{\text{f}} = 0.30$
and $\alpha_{\text{a}} = 0.10$.
We assign equal weights to heir identification, share prediction, and final distribution because they contain the main information needed to solve the case. They evaluate the model’s ability to identify the eligible heirs, assign the correct shares, and produce the final distribution. In contrast, the adjustment step receives a smaller weight because it only checks whether the case involves \textit{ʿawl}, \textit{radd}, or neither.
\section{Experiments and Results} \label{exp}

This section describes the experimental setup, evaluated models, and the
results obtained on the MAWARITH dataset.
\begin{figure*}[t]
  \centering
  \includegraphics[width=\textwidth,height=0.42\textheight,keepaspectratio]{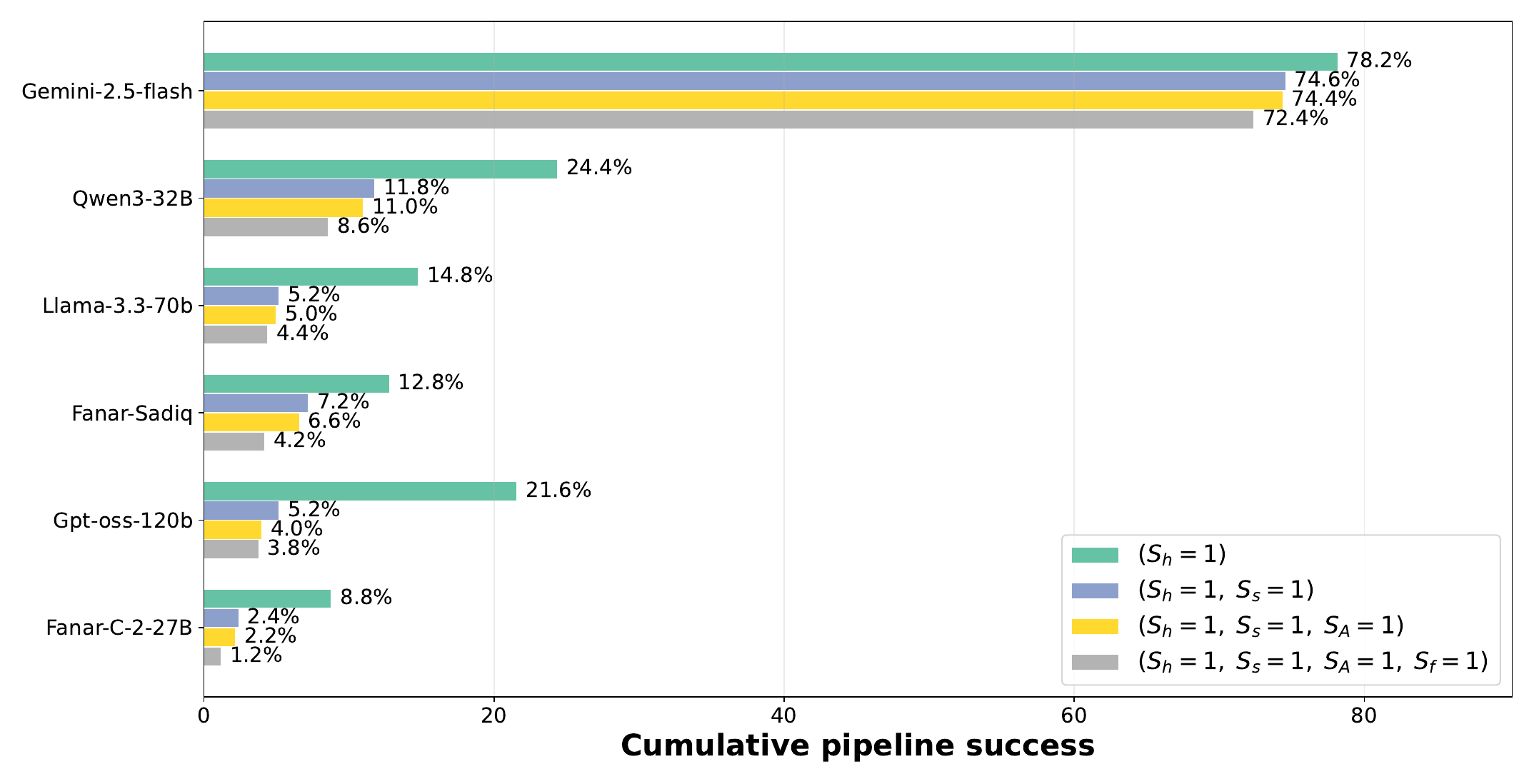}
  \caption{Cumulative pipeline success rates across stages (Step1--Step4) for each model.}
  \label{fig:steps_pipeline_success}
\end{figure*}

\subsection{Experimental Setup}
We evaluate six LLMs in a zero-shot setting using Arabic prompts, without any task-specific fine-tuning. The prompt is provided in ~\ref{app:reasoning_prompt}. It specifies the inheritance scenario, the required reasoning steps, and the target output schema, and it enforces adherence to the formal rules of Islamic inheritance law, enabling a fair and controlled comparison across models. The evaluated models include Fanar\footnote{\url{https://api.fanar.qa/docs\#description/introduction}} (an Arabic-centric LLM, evaluated in both its general-purpose variant, \textit{Fanar-C-2-27B}, and its Islamic-specialized variant, \textit{Fanar-Sadiq}), open-source multilingual models (LLaMA~3~70B\footnote{\url{https://huggingface.co/meta-llama/Meta-Llama-3-70B}} and GPT-OSS~120B\footnote{\url{https://huggingface.co/openai/gpt-oss-120b}}), a multilingual reasoning model (Qwen3-32B\footnote{\url{https://huggingface.co/Qwen/Qwen3-32B}}), and a commercial model (Gemini-2.5-flash). Model sizes range from 27B to over 120B parameters; however, the exact parameter count of Gemini is not publicly disclosed. We obtained predictions via the Fanar API for both Fanar models and the Groq API\footnote{\url{https://console.groq.com/}} for LLaMA and GPT-OSS. For Qwen3-32B, we ran inference locally on a server with three GPUs (144GB total memory). Across all models, we use a context window of $4096$ tokens and set the decoding temperature to $0.1$. For each instance, the model is required to produce outputs in the structured format described in Section~\ref{struc}, which enables fine-grained evaluation and error analysis by comparing intermediate decisions against the gold reference.
\\
For validation, we randomly selected 200 instances from the training corpus and held them out as a validation set. All models were then evaluated on the same benchmark split, consisting of 200 validation instances and 500 test instances.

\subsection{Overall Performance and Reasoning Analysis}
Table~\ref{tab:main_results} reports the overall MIR-E scores obtained by all evaluated models, including their performance across the main reasoning steps (heir identification, share assignment, and final distribution accuracy) on both the validation and test sets.
A similar performance pattern is observed on both the validation and test sets, with Gemini clearly outperforming the other models and achieving a MIR-E score close to 90\%. In contrast, the open-weight models obtain overall MIR-E scores below 45\%. Among them, Qwen3-32B achieves the strongest performance, followed by Fanar-Sadiq.
\\
Although Qwen3-32B, Fanar-Sadiq, and GPT-OSS-120B are capable of generating multi-step reasoning, they often make early-stage errors (e.g., incorrect heir identification or blocking decisions), which then propagate and invalidate the final allocation. This behavior is expected in inheritance reasoning, where the steps are strongly interdependent: each heir’s share depends on the presence or absence of other heirs, their assigned shares, and the correct application of arithmetic and normalization rules. As a result, intermediate mistakes can cascade into a fully incorrect solution.
\begin{table*}[t]
  \centering
  \footnotesize
  \setlength{\tabcolsep}{4pt}
  \renewcommand{\arraystretch}{1.05}
  \begin{tabular}{@{}ll|cccccccc@{}}
    \toprule
    & & \multicolumn{4}{c}{\textbf{Validation}} & \multicolumn{4}{c}{\textbf{Test}} \\
    \cmidrule(lr){3-6} \cmidrule(lr){7-10}
    \textbf{Model} &
    \textbf{} &
    $S_{\mathrm{heirs}}$ &
    $S_{\mathrm{shares}}$ &
    $S_{\mathrm{norm}}$ &
    \cellcolor{gray!20}\textbf{MIR-E} &
    $S_{\mathrm{heirs}}$ &
    $S_{\mathrm{shares}}$ &
    $S_{\mathrm{norm}}$ &
    \cellcolor{gray!20}\textbf{MIR-E} \\
    \midrule
    C-ML-R  & Gemini-2.5-flash & 0.965 & 0.899 & 0.892 & \cellcolor{gray!20}\textbf{0.918} & 0.945 & 0.929 & 0.894 & \cellcolor{gray!20}\textbf{0.901} \\
    OS-ML-R & Qwen3-32B        & 0.761 & 0.439 & 0.237 & \cellcolor{gray!20}\textbf{0.438} & 0.690 & 0.446 & 0.265 & \cellcolor{gray!20}\textbf{0.437} \\
    OS-ML   & GPT-oss-120B     & 0.692 & 0.311 & 0.273 & \cellcolor{gray!20}\textbf{0.389} & 0.693 & 0.327 & 0.271 & \cellcolor{gray!20}\textbf{0.391} \\
    OS-ML   & Llama-3.3        & 0.676 & 0.321 & 0.148 & \cellcolor{gray!20}\textbf{0.345} & 0.648 & 0.403 & 0.215 & \cellcolor{gray!20}\textbf{0.390} \\
    OS-A    & Fanar-Sadiq      & 0.655 & 0.443 & 0.204 & \cellcolor{gray!20}\textbf{0.393} & 0.621 & 0.367 & 0.204 & \cellcolor{gray!20}\textbf{0.368} \\
    OS-A    & Fanar-C-2-27B    & 0.645 & 0.340 & 0.176 & \cellcolor{gray!20}\textbf{0.351} & 0.584 & 0.314 & 0.178 & \cellcolor{gray!20}\textbf{0.328} \\
    \bottomrule
  \end{tabular}
  \caption{
    (C: commercial, OS: open-source, ML: multilingual, A: Arabic, R: reasoning.)
    Component-wise MIR-E scores on the validation (200 cases) and test (500 cases) sets.
    Validation results are used for model selection, while all comparisons and conclusions are based on the held-out test set.\\
  }
  \label{tab:main_results}
\end{table*}
Figure~\ref{fig:steps_pipeline_success}  shows the \emph{cumulative} success rate across inheritance reasoning pipeline, where each stage is evaluated conditionally on all preceding stages being correct (Step1: heir identification and blocking; Step2: share assignment; Step3: adjustment; Step4:  final allocation).  We observe that, \emph{except for} Gemini, the models struggle to correctly identify eligible heirs and exclude blocked heirs: Gemini  achieves  a  success rate of 78.2\%, whereas all other models remain below 25\% at this stage. It also appears that errors propagate: early mistakes significantly  reduce the number of cases that can succeed at later stages.  All open-weight models experience their largest degradation between Step1 and Step2, highlighting share assignment as the main bottleneck under fixed heirs (\textit{e.g.}, Qwen3-32B drops from 24.4\% to 11.8\%).
This performance gap may be partially explained by differences in \textit{model scale} and \textit{training data}.  Gemini benefits from a larger scale and proprietary training resources. Model scale and training data diversity likely contribute to more reliable multi-step reasoning under structured legal constraints.

\section{Analysis of Errors Across Reasoning Stages}
\label{analysis}

To better understand model weaknesses across the inheritance reasoning pipeline, we perform a fine-grained, stage-level error analysis. Our findings show that errors extend beyond the application of Islamic inheritance rules to earlier linguistic parsing steps. Indeed, all models exhibit linguistic parsing errors when extracting heirs from Arabic problem statements. For example, consider the following case: \foreignlanguage{arabic}{مات وترك: أخوين شقيقين وخمس أخوات شقيقات وأب الأب وأربع بنات ابن ابن. ما هو نصيب كل وارث؟}
(“A person died leaving: two full brothers, five full sisters, the paternal grandfather (the deceased’s father’s father), and four great-granddaughters through the son’s line (daughters of the son’s son). What is the share of each heir?”). In this example, some models fail to correctly interpret the compound kinship expression \foreignlanguage{arabic}{أربع بنات ابن ابن} (\textit{i.e.}, four daughters of a son's son). Instead of treating it as a single heir type with an associated count, the models incorrectly decompose it into two separate heirs (e.g., \foreignlanguage{arabic}{بنات ابن} (\textit{i.e.}, four daughters of a son) plus \foreignlanguage{arabic}{ابن} (\textit{i.e.}, a son)), effectively splitting one heir into two and introducing spurious heirs.
\\
Moreover, models sometimes modify the heir type names even when explicitly instructed to preserve them exactly as provided. For instance, the model may replace \foreignlanguage{arabic}{``أب الأب''} with the equivalent term \foreignlanguage{arabic}{``الجد''}, despite the prompt explicitly requiring the model to keep the original heir names without introducing synonyms or alternative expressions. This shows that LLMs often rephrase the heirs, which can bias the evaluation if it relies on exact string matching and does not account for semantic equivalence.
\\
We observe that this error type is relatively rare for Gemini, yet frequent in the other models. Even with an explicit prompt that provides a complete list of heir types and requires the model to select heirs only from this predefined list, such errors still occur, showing that linguistic understanding remains a key bottleneck in the reasoning pipeline.
\\
To focus the evaluation on the models' ability to apply Islamic inheritance knowledge rather than on their ability to parse complex Arabic descriptions, we explicitly present the heirs in a structured form. Instead of requiring the model to identify heirs only from the full narrative description, we provide the extracted heirs as a separate list, keeping both the heir type and the count in natural language (e.g., \foreignlanguage{arabic}{[أخوين شقيقين، خمس أخوات شقيقات، أب الأب، أربع بنات ابن ابن]}).
\\
This design reduces linguistic ambiguity and allows the evaluation to primarily assess whether the model correctly applies inheritance rules and computes the appropriate shares. In practice, many models, especially smaller ones, frequently make errors at the heir-extraction stage.  By controlling for linguistic parsing errors, the benchmark does not primarily evaluate the models' ability to parse complex Arabic heir expressions, but instead focuses on their ability to apply Islamic inheritance rules once the heirs are explicitly provided. This design therefore provides a more accurate measure of the models' reasoning capabilities in Islamic inheritance law.

\subsection{Heir Identification Errors}

Heir identification is the foundational step of inheritance reasoning, since all subsequent stages depend on predicting the correct set of eligible heirs. We distinguish three types of structural errors: \textit{\textbf{(i) False Blocking (FB):}} where an eligible heir is incorrectly excluded, which decreases the number of beneficiaries; \textit{\textbf{(ii) False Eligibility (FE):}} where a non-entitled heir is incorrectly included, increasing the number of beneficiaries; and \textit{\textbf{(iii) Count Mismatch Errors (ME):}} where the model predicts wrong heir counts due to misreading Arabic expressions in the question (e.g., numbers and plural forms), by adding or reducing the number of heirs, which then affects the shares assigned to the other heirs.

\begin{table}[!ht]
\centering
\small
\begin{tabular}{lcccc}
\toprule
Model & FB & FE & ME & Total\_err \\
\midrule
Gemini-2.5-flash & 77  & 66  & 0 & 143  \\
Fanar-Sadiq      & 285 & 456 & 0 & 741  \\
GPT-OSS-120B     & 415 & 327 & 1 & 743  \\
LLaMA-3.3-70B    & 287 & 619 & 1 & 907  \\
Fanar-C-2-27B    & 168 & 937 & 0 & 1105 \\
\midrule
\textbf{Total} & \textbf{1232} & \textbf{2405} & \textbf{2} & \textbf{3639} \\
\bottomrule
\end{tabular}
\caption{Structural heir-identification errors on the test corpus. FB: false blocking; FE: false eligibility; ME: count mismatch errors; Total\_err: total errors.}
\label{tab:structural_errors}
\end{table}

Table~\ref{tab:structural_errors} reports the distribution of heir-identification errors (\textit{FB/FE/ME}) across models. The results highlight large differences in models' ability to apply basic inheritance rules. Gemini-2.5-flash achieves a score of 0.94 on this step, while the other models do not exceed 0.70. We also observe that GPT-OSS-120B performs even better than the Arabic model Fanar. Fanar-C-2-27B often makes false-eligibility errors: it adds extra heirs who should not inherit because they are blocked. In contrast, its Islamic-adapted version (with RAG) selects eligible heirs more accurately, because it retrieves the relevant blocking rules from trusted inheritance references (e.g., classical inheritance books or close fatwa-style rulings) and uses them to guide the eligibility decision.

Across most systems, false eligibility is the dominant failure mode, suggesting that models often add heirs to the heir list even though they have no right to inherit under Islamic inheritance law. This is followed by false blocking errors, where the model incorrectly excludes heirs who are entitled to inherit. Count mismatch errors are rare and appear mostly in open-source multilingual models (e.g., LLaMA and GPT-OSS). These errors are less related to inheritance knowledge and are more likely due to Arabic number understanding (numbers written in words) and plural morphology. For example, \foreignlanguage{arabic}{خمس بنات ابن} (\emph{five son's daughters}) should be interpreted as (\foreignlanguage{arabic}{بنت ابن} $\times$ 5), but the model predicted (\foreignlanguage{arabic}{بنت ابن} $\times$ 3). In this case, the heir category is correct, but the numeric count is wrong. We consider this an isolated error, since the same heir type with the correct count is predicted correctly in many other instances.
For example, in the case: \foreignlanguage{arabic}{``مات وترك: أخوين شقيقين وخمس أخوات شقيقات وأب الأب وأربع بنات ابن ابن. ما هو نصيب كل وارث؟''}, some models fail to correctly interpret the compound kinship expression \foreignlanguage{arabic}{``أربع بنات ابن ابن''} (\textit{i.e.}, four great-granddaughters through the son's son). Instead of treating it as a single heir type with multiplicity, the model incorrectly decomposes it into two separate heirs (e.g., \foreignlanguage{arabic}{بنات ابن} $\times 4$ plus \foreignlanguage{arabic}{ابن} $\times 1$), effectively splitting one heir into two and introducing spurious heirs. Although our prompt explicitly guides the model to extract the complete heir list carefully before applying inheritance rules, such errors still occur, showing that linguistic understanding remains a key bottleneck in the reasoning pipeline.\\
\begin{figure*}[t]
    \centering
    \includegraphics[width=\linewidth]{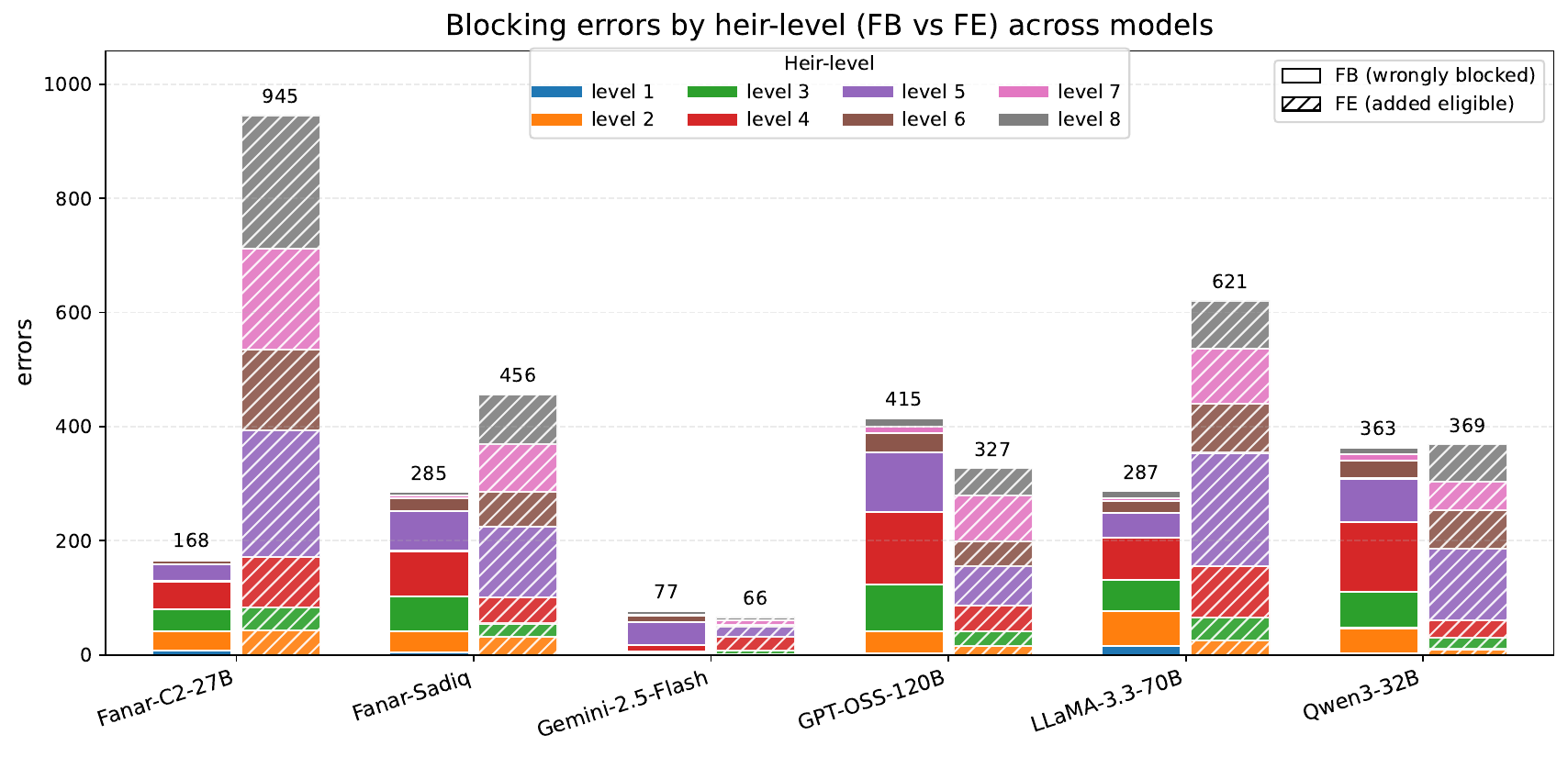}
    \caption{Blocking errors by genealogical heir level across models. For each model, the left bar shows false blocking (FB: wrongly blocked eligible heirs) and the right bar shows false eligibility (FE: added eligible heirs). Colors indicate heir levels (1--8), as defined in ~\ref{appendix/appendix_heirs}.}
    \label{fig_blocking_by_level}
\end{figure*}
To study models' inheritance knowledge in more detail, we analyse \textit{false blocking} (FB) and \textit{false eligibility} (FE) by genealogical heir level. Figure~\ref{fig_blocking_by_level} shows heir-identification errors across Levels~1--8, separating \textit{false blocking} (FB) and \textit{false eligibility} (FE). The heirs included in each level are listed in ~\ref{appendix/appendix_heirs}. Levels~1--2 correspond to close family (spouse, parents, and children), while Levels~5--8 correspond to more distant relatives (such as uncles and cousins).
\\
We observe that most models (except \textit{Gemini}) often add distant heirs who are not eligible when closer heirs are present. For example, LLaMA sometimes includes brothers and uncles even when a son is present, even though a son blocks these collateral relatives under \textit{mawārīth} rules. Such errors are relatively rare for close relatives (Levels~1--3), who are generally eligible, but become more frequent in the middle tiers and among the most distant levels. At the same time, models tend to make the opposite mistake less often: the most distant heirs are the least frequently \emph{wrongly blocked}, which is intuitive since they are typically blocked when closer heirs are present.
\\
These results can be explained by the fact that these models were not trained specifically for inheritance-law reasoning. In practice, they often approximate an answer by \textit{(i)} matching the query to the most similar case seen during pretraining (\textit{e.g.}, in fatwas) and reusing its outcome, or \textit{(ii)} applying isolated rules extracted from Islamic sources without ensuring global consistency across the full set of constraints. They may have some general knowledge about inheritance terms, but this knowledge is often insufficient to apply the full set of rules consistently across cases.

\subsection{Share Errors}
\label{sec:share_errors}
Share errors can be produced by two factors: \textit{(i)} limited knowledge of the jurisprudential rules that determine each heir's fraction, and \textit{(ii)} error propagation from the identification step. Indeed, \textit{mawārīth} shares depend on the list of eligible heirs, and any change in this list directly affects the share assigned to each heir. We analyse share errors by \textit{keeping only test examples where the selection of eligible heirs is correct}, and then computing the mean share score on this subset. Table~\ref{tab:shares_hb1} compares models on share computation after removing error propagation from the heir-identification step. We keep only test cases where heir identification is perfectly correct ($S_{\text{heirs}}=1$) and compute the mean share score $S_{\text{shares}}$ on this subset.
\begin{table}[h]
\centering
\begin{tabular}{lrr}
\hline
Model & $\#(S_h{=}1)$ & $\#(S_s \mid S_h{=}1)$ \\
\hline
Gemini-2.5-flash & 390 & 0.977 \\
Fanar-Sadiq      & 64  & 0.727 \\
Llama-3.3-70b    & 74  & 0.612 \\
Fanar-C-2-27B    & 44  & 0.542 \\
GPT-oss-120b     & 108 & 0.526 \\
\hline
\end{tabular}
\caption{Share-allocation score conditioned on correct heir identification.\\}
\label{tab:shares_hb1}
\end{table}
\textit{Gemini-2.5-flash} achieves a very high share score ($S_{\text{shares}}=0.977$) on 390 examples, showing strong rule-based share computation when the eligible-heir set is correct. In contrast, the other models have much smaller subsets (from $N_{S_{\text{heirs}}=1}=44$ to 108) and their share scores are lower ($S_{\text{shares}}=0.526$--$0.727$). This indicates that, beyond \textit{FB/FE} propagation, these models still make share-calculation errors even when the heir set is correct, reflecting weaker coverage of \textit{mawārīth} allocation rules.

For example, in the following inheritance case: \foreignlanguage{arabic}{مات وترك: أربع أخوات لأم، وأم، وأم الأب، وثلاثة أبناء أخ لأب، وأب الأب، وأب. ما هو نصيب كل وريث؟}, the heirs mentioned include the mother (\foreignlanguage{arabic}{الأم}), the father (\foreignlanguage{arabic}{الأب}), and four maternal half-sisters of the decedent (\foreignlanguage{arabic}{أربع أخوات لأم}). Even when the model correctly identifies the eligible heirs (only the parents inherit, since the father excludes the sisters), it may still incorrectly assign the mother a share of $1/3$ by citing:
\foreignlanguage{arabic}{``فَإِن لَّمْ يَكُن لَّهُ وَلَدٌ وَوَرِثَهُ أَبَوَاهُ فَلِأُمِّهِ الثُّلُثُ'' } \hfill (Q~4:11).
However, the same verse immediately states an explicit exception:
\foreignlanguage{arabic}{``فَإِن كَانَ لَهُ إِخْوَةٌ فَلِأُمِّهِ السُّدُسُ''} \hfill (Q~4:11). 
Since \foreignlanguage{arabic}{أخوات لأم} are siblings of the decedent, the condition \foreignlanguage{arabic}{``له إخوة''} is satisfied, which reduces the mother's share to $1/6$ under the majority view (\foreignlanguage{arabic}{الجمهور}). Therefore, the correct allocation is mother $1/6$ and father $5/6$ (the remainder by \foreignlanguage{arabic}{التعصيب}), whereas the model predicts mother $=1/3$ and father $2/3$. This illustrates that some models exhibit \emph{rule-knowledge gaps} in applying Qur'anic conditional clauses, even when the eligible-heir set is correctly determined.

\subsection{Adjustment Errors}

We evaluate the adjustment step only on instances where blocking decisions and share assignment are correct (\textit{i.e.}, $B{=}1$ and $S_s{=}1$). This restriction is important because the adjustment (\textit{ʿawl/radd}) outcome is determined by the assigned fractions and their sum: \textit{ʿawl} applies when the total exceeds~1, while \textit{radd} applies when the total is below~1 and no residuary heir (\textit{ʿaṣabah}) absorbs the remainder. If heirs are incorrectly blocked/unblocked or shares are incorrect (wrong fractions, missing/extra heirs), the adjustment decision is not interpretable.
\begin{table*}[t]
\centering
\setlength{\tabcolsep}{4pt}
\begin{tabular}{lrrr}
\toprule
{Model} &
{\small $\#(S_h{=}1,S_s{=}1)$} &
{\small $\#(S_a{=}1 \mid S_h{=}1,S_s{=}1)$} &
{\small $\#(S_a{=}0 \mid S_h{=}1,S_s{=}1)$ (\%)} \\
\midrule
Gemini-2.5-flash & 373 & 372 & 1 (0.27) \\
Qwen3-32B        & 59  & 55  & 4 (6.78) \\
Fanar-Sadiq      & 36  & 33  & 3 (8.33) \\
GPT-oss-120b     & 26  & 20  & 6 (23.08) \\
Llama-3.3-70b    & 26  & 25  & 1 (3.85) \\
Fanar-C-2-27B    & 12  & 11  & 1 (8.33) \\
\bottomrule
\end{tabular}
\caption{Adjustment accuracy conditioned on correct previous steps (heir identification $S_h{=}1$ and share assignment $S_s{=}1$). We report the subset size $\#(S_h{=}1,S_s{=}1)$ and the number of correct/incorrect adjustment outcomes ($S_a{=}1/0$) within this subset.}
\label{tab:adjustment_given_Ss1_B1}
\end{table*}
\\
Table~\ref{tab:adjustment_given_Ss1_B1} shows that Gemini-2.5-flash makes only one error out of 373 adjustment-evaluable instances (0.27\%). Most other models make only a few adjustment mistakes (3.85--8.33\%), whereas GPT-OSS-120B exhibits a higher error rate (23.08\%). This means that, once earlier steps are correct, the adjustment step is generally reliable for most models.
\\
A typical adjustment error is a confusion between \emph{remainder handling} and \textit{radd}. When the sum of fixed shares is below 1, the remainder should be absorbed by a residuary heir (taṣīl) if one exists; otherwise, it should be redistributed through \textit{radd} rules. In several cases, models label these instances as \textit{radd} (or omit the adjustment step) without checking for a residuary heir, which results in an incorrect adjustment decision even when the fixed fractions are otherwise correct.
\\
A second error type occurs when the model skips the adjustment step, failing to determine whether the case requires \textit{ʿawl} or \textit{radd}, and instead proceeding directly from share assignment to the final allocation. For example, with a wife ($1/8$), a mother ($1/6$), and two daughters ($2/3$), the fixed shares sum to $23/24$ ($<1$), so the case requires handling a remainder (\textit{radd}) rather than applying \textit{ʿawl}. However, some models omit this check and finalize the distribution without accounting for the remainder. In other cases, they wrongly treat the case as \textit{ʿawl} even though the total does not exceed one, effectively bypassing the adjustment step.

\subsection{Final Allocation Errors}

This category captures \emph{computational} failures rather than lack of knowledge of \textit{mawārīth} rules. As shown in Figure~\ref{fig:steps_pipeline_success}, such errors are relatively rare. They mainly occur at the last stage of the pipeline, \emph{final allocation}. In many cases, models correctly identify that an heir should receive \foreignlanguage{arabic}{باقي التركة} (the remainder), but fail to \emph{materialize} this remainder into explicit numeric final shares. In some cases, models miscompute the remaining share after allocating the fixed fractions and/or misconvert it into percentages, leading to incorrect final allocations.
\\
It is important to note that in our dataset, \textit{ʿawl} and \textit{radd} require recomputing the heirs' shares. However, in our evaluation we only assess whether a case requires adjustment and which type applies (\textit{ʿawl} or \textit{radd}). This strategy is adopted because most instances in our corpus are simple cases without adjustment, where an adjusted-shares output is not applicable; scoring adjusted shares in such cases would introduce inconsistencies in the evaluation metric. Moreover, when adjustment is required, the adjusted shares largely coincide with the final allocation, so any error in the recomputed shares would directly propagate to the final allocation step. We also observe a less common error where models report group-level shares rather than per-individual allocations (e.g., one value for ``three brothers'' rather than one value per brother).

\section{Conclusion}
\label{sec:conclusion}

This paper introduced \textsc{MAWARITH}, a dataset and benchmark for evaluating \emph{legal reasoning} in Arabic Islamic inheritance law. 
\textsc{MAWARITH} is designed to support training for structured reasoning by breaking inheritance cases into explicit step-by-step stages.
 Each stage is  explained and justified with relevant Islamic legal texts, and requires accurate jurisprudential knowledge and strict rule application: \textit{(i)} identifying eligible heirs and blocking relations, \textit{(ii)} assigning fixed shares and residuary portions, and \textit{(iii)} producing a final distribution. To support fine-grained and interpretable evaluation, we propose \textit{MIR-E}, a metric that breaks down performance into step-level subscores aligned with the inheritance reasoning pipeline, enabling fine-grained error analysis.
Our evaluation highlights a clear gap between commercial  and open-weight models. 
Although some models can generate convincing explanations, they often fail at early structural steps (e.g., missing or hallucinated heirs, incorrect blocking). 
These early errors propagate to later steps, leading to inconsistent share assignments and invalid final distributions. We hypothesize that a primary cause is limited exposure to inheritance-law knowledge during training, since most training data is generic and insufficient to solve inheritance problems end-to-end. We expect \textsc{MAWARITH} to be an important resource for building reasoning and Islamic-domain LLMs that can solve inheritance cases with detailed, step-by-step reasoning.
\\
Future work will explore reinforcement learning approaches, including Process Reward Models, to better align model reasoning with legal constraints. 
By providing step-level feedback, these methods can guide models toward valid reasoning trajectories and reduce error propagation in multi-step legal reasoning. We will also extend the corpus to cover more complex remaining inheritance cases, including pregnancy-related cases, multiple deaths, the missing person (\textit{mafqūd}), and the intersex heir (\textit{khunthā}).



\bibliography{custom}

\begin{thebibliography}{34}
\expandafter\ifx\csname natexlab\endcsname\relax\def\natexlab#1{#1}\fi
\providecommand{\url}[1]{\texttt{#1}}
\providecommand{\href}[2]{#2}
\providecommand{\path}[1]{#1}
\providecommand{\DOIprefix}{doi:}
\providecommand{\ArXivprefix}{arXiv:}
\providecommand{\URLprefix}{URL: }
\providecommand{\Pubmedprefix}{pmid:}
\providecommand{\doi}[1]{\href{http://dx.doi.org/#1}{\path{#1}}}
\providecommand{\Pubmed}[1]{\href{pmid:#1}{\path{#1}}}
\providecommand{\bibinfo}[2]{#2}
\ifx\xfnm\relax \def\xfnm[#1]{\unskip,\space#1}\fi
\bibitem[{Cobbe et~al.(2021)Cobbe, Kosaraju, Bavarian, Chen, Jun, Kaiser,
  Plappert, Tworek, Hilton, Nakano et~al.}]{cobbe2021training}
\bibinfo{author}{K.~Cobbe}, \bibinfo{author}{V.~Kosaraju},
  \bibinfo{author}{M.~Bavarian}, \bibinfo{author}{M.~Chen},
  \bibinfo{author}{H.~Jun}, \bibinfo{author}{L.~Kaiser},
  \bibinfo{author}{M.~Plappert}, \bibinfo{author}{J.~Tworek},
  \bibinfo{author}{J.~Hilton}, \bibinfo{author}{R.~Nakano}, et~al.,
\newblock \bibinfo{title}{Training verifiers to solve math word problems},
\newblock \bibinfo{journal}{arXiv preprint arXiv:2110.14168}
  (\bibinfo{year}{2021}).
\bibitem[{Hendrycks et~al.(2021)Hendrycks, Burns, Kadavath, Arora, Basart,
  Tang, Song, and Steinhardt}]{hendrycks2021math}
\bibinfo{author}{D.~Hendrycks}, \bibinfo{author}{C.~Burns},
  \bibinfo{author}{S.~Kadavath}, \bibinfo{author}{A.~Arora},
  \bibinfo{author}{S.~Basart}, \bibinfo{author}{E.~Tang},
  \bibinfo{author}{D.~Song}, \bibinfo{author}{J.~Steinhardt},
\newblock \bibinfo{title}{Measuring mathematical problem solving with the math
  dataset},
\newblock \bibinfo{journal}{arXiv preprint arXiv:2103.03874}
  (\bibinfo{year}{2021}).
\bibitem[{Suzgun et~al.(2022)Suzgun, Scales, Sch{\"a}rli, Gehrmann, Tay, Chung,
  Chowdhery, Le, Chi, Zhou, and Wei}]{suzgun2022bbh}
\bibinfo{author}{M.~Suzgun}, \bibinfo{author}{N.~Scales},
  \bibinfo{author}{N.~Sch{\"a}rli}, \bibinfo{author}{S.~Gehrmann},
  \bibinfo{author}{Y.~Tay}, \bibinfo{author}{H.~W. Chung},
  \bibinfo{author}{A.~Chowdhery}, \bibinfo{author}{Q.~V. Le},
  \bibinfo{author}{E.~H. Chi}, \bibinfo{author}{D.~Zhou},
  \bibinfo{author}{J.~Wei},
\newblock \bibinfo{title}{Challenging {BIG}-{Bench} tasks and whether
  chain-of-thought can solve them},
\newblock \bibinfo{journal}{arXiv preprint arXiv:2210.09261}
  (\bibinfo{year}{2022}).
\bibitem[{Chalkidis et~al.(2022)Chalkidis, Jana, Hartung, Bommarito,
  Androutsopoulos, Katz, and Aletras}]{chalkidis2022lexglue}
\bibinfo{author}{I.~Chalkidis}, \bibinfo{author}{A.~Jana},
  \bibinfo{author}{D.~Hartung}, \bibinfo{author}{M.~Bommarito},
  \bibinfo{author}{I.~Androutsopoulos}, \bibinfo{author}{D.~M. Katz},
  \bibinfo{author}{N.~Aletras},
\newblock \bibinfo{title}{Lexglue: A benchmark dataset for legal language
  understanding in english},
\newblock in: \bibinfo{booktitle}{Proceedings of the 60th Annual Meeting of the
  Association for Computational Linguistics (ACL)}, \bibinfo{year}{2022}.
\bibitem[{Niklaus et~al.(2023)Niklaus, Matoshi, Rani, Galassi, St{\"u}rmer, and
  Chalkidis}]{niklaus2023lextreme}
\bibinfo{author}{J.~Niklaus}, \bibinfo{author}{V.~Matoshi},
  \bibinfo{author}{P.~Rani}, \bibinfo{author}{A.~Galassi},
  \bibinfo{author}{M.~St{\"u}rmer}, \bibinfo{author}{I.~Chalkidis},
\newblock \bibinfo{title}{Lextreme: A multi-lingual and multi-task benchmark
  for the legal domain},
\newblock in: \bibinfo{booktitle}{Findings of EMNLP}, \bibinfo{year}{2023}.
\bibitem[{Guha et~al.(2023)Guha, Nyarko, Ho, R{\'e}, Chilton
  et~al.}]{guha2023legalbench}
\bibinfo{author}{N.~Guha}, \bibinfo{author}{J.~Nyarko}, \bibinfo{author}{D.~E.
  Ho}, \bibinfo{author}{C.~R{\'e}}, \bibinfo{author}{A.~Chilton}, et~al.,
\newblock \bibinfo{title}{Legalbench: A collaboratively built benchmark for
  measuring legal reasoning in large language models},
\newblock \bibinfo{journal}{arXiv preprint arXiv:2308.11462}
  (\bibinfo{year}{2023}).
\bibitem[{Pal et~al.(2022)Pal, Umapathi, and Sankarasubbu}]{pal2022medmcqa}
\bibinfo{author}{A.~Pal}, \bibinfo{author}{L.~K. Umapathi},
  \bibinfo{author}{M.~Sankarasubbu},
\newblock \bibinfo{title}{Medmcqa: A large-scale multi-subject multi-choice
  dataset for medical domain question answering},
\newblock in: \bibinfo{booktitle}{Proceedings of the Conference on Health,
  Inference, and Learning (CHIL)}, \bibinfo{year}{2022}.
\bibitem[{Kim et~al.(2024)Kim, Wu, Abdulle, and Wu}]{kim2024medexqa}
\bibinfo{author}{Y.~Kim}, \bibinfo{author}{J.~Wu},
  \bibinfo{author}{Y.~Abdulle}, \bibinfo{author}{H.~Wu},
\newblock \bibinfo{title}{Medexqa: Medical question answering benchmark with
  multiple explanations},
\newblock in: \bibinfo{booktitle}{Proceedings of the Workshop on Biomedical
  Language Processing (BioNLP)}, \bibinfo{year}{2024}.
\bibitem[{Bhatia et~al.(2026)Bhatia, Mubarak, Jarrar, Mikros, Zaraket,
  Alhirthani, Al-Khatib, Cochrane, Darwish, Yahiaoui
  et~al.}]{bhatia2026ragagenticragfaithful}
\bibinfo{author}{G.~Bhatia}, \bibinfo{author}{H.~Mubarak},
  \bibinfo{author}{M.~Jarrar}, \bibinfo{author}{G.~Mikros},
  \bibinfo{author}{F.~Zaraket}, \bibinfo{author}{M.~Alhirthani},
  \bibinfo{author}{M.~Al-Khatib}, \bibinfo{author}{L.~Cochrane},
  \bibinfo{author}{K.~Darwish}, \bibinfo{author}{R.~Yahiaoui}, et~al.,
\newblock \bibinfo{title}{From {RAG} to agentic {RAG} for faithful islamic
  question answering},
\newblock \bibinfo{journal}{arXiv preprint arXiv:2601.07528}
  (\bibinfo{year}{2026}).
\bibitem[{Mubarak et~al.(2025)Mubarak, Malhas, Mansour, Mohamed, Fawzi,
  Hawasly, Elsayed, Darwish, and Magdy}]{mubarak-etal-2025-islamiceval}
\bibinfo{author}{H.~Mubarak}, \bibinfo{author}{R.~Malhas},
  \bibinfo{author}{W.~Mansour}, \bibinfo{author}{A.~Mohamed},
  \bibinfo{author}{M.~Fawzi}, \bibinfo{author}{M.~Hawasly},
  \bibinfo{author}{T.~Elsayed}, \bibinfo{author}{K.~M. Darwish},
  \bibinfo{author}{W.~Magdy},
\newblock \bibinfo{title}{{I}slamic{E}val 2025: The first shared task of
  capturing {LLM}s hallucination in islamic content},
\newblock in: \bibinfo{editor}{K.~Darwish}, \bibinfo{editor}{A.~Ali},
  \bibinfo{editor}{I.~Abu~Farha}, \bibinfo{editor}{S.~Touileb},
  \bibinfo{editor}{I.~Zitouni}, \bibinfo{editor}{A.~Abdelali},
  \bibinfo{editor}{S.~Al-Ghamdi}, \bibinfo{editor}{S.~Alkhereyf},
  \bibinfo{editor}{W.~Zaghouani}, \bibinfo{editor}{S.~Khalifa},
  \bibinfo{editor}{B.~AlKhamissi}, \bibinfo{editor}{R.~Almatham},
  \bibinfo{editor}{I.~Hamed}, \bibinfo{editor}{Z.~Alyafeai},
  \bibinfo{editor}{A.~Alowisheq}, \bibinfo{editor}{G.~Inoue},
  \bibinfo{editor}{K.~Mrini}, \bibinfo{editor}{W.~Alshammari} (Eds.),
  \bibinfo{booktitle}{Proceedings of The Third Arabic Natural Language
  Processing Conference: Shared Tasks}, \bibinfo{publisher}{Association for
  Computational Linguistics}, \bibinfo{address}{Suzhou, China},
  \bibinfo{year}{2025}, pp. \bibinfo{pages}{480--493}.
  \DOIprefix\doi{10.18653/v1/2025.arabicnlp-sharedtasks.67}.
\bibitem[{Xuan~Phuc and
  {\DJ}ặng~V{\u{a}}n(2025)}]{xuan-phuc-dang-van-2025-puxai}
\bibinfo{author}{N.~Xuan~Phuc}, \bibinfo{author}{T.~{\DJ}ặng~V{\u{a}}n},
\newblock \bibinfo{title}{{P}ux{AI} at {QIAS} 2025: Multi-agent
  retrieval-augmented generation for islamic inheritance and knowledge
  reasoning},
\newblock in: \bibinfo{editor}{K.~Darwish}, \bibinfo{editor}{A.~Ali},
  \bibinfo{editor}{I.~Abu~Farha}, \bibinfo{editor}{S.~Touileb},
  \bibinfo{editor}{I.~Zitouni}, \bibinfo{editor}{A.~Abdelali},
  \bibinfo{editor}{S.~Al-Ghamdi}, \bibinfo{editor}{S.~Alkhereyf},
  \bibinfo{editor}{W.~Zaghouani}, \bibinfo{editor}{S.~Khalifa},
  \bibinfo{editor}{B.~AlKhamissi}, \bibinfo{editor}{R.~Almatham},
  \bibinfo{editor}{I.~Hamed}, \bibinfo{editor}{Z.~Alyafeai},
  \bibinfo{editor}{A.~Alowisheq}, \bibinfo{editor}{G.~Inoue},
  \bibinfo{editor}{K.~Mrini}, \bibinfo{editor}{W.~Alshammari} (Eds.),
  \bibinfo{booktitle}{Proceedings of The Third Arabic Natural Language
  Processing Conference: Shared Tasks}, \bibinfo{publisher}{Association for
  Computational Linguistics}, \bibinfo{address}{Suzhou, China},
  \bibinfo{year}{2025}, pp. \bibinfo{pages}{905--913}.
  \DOIprefix\doi{10.18653/v1/2025.arabicnlp-sharedtasks.125}.
\bibitem[{Bouchekif et~al.(2025)Bouchekif, Rashwani, Sbahi, Gaben, Al-Khatib,
  and Ghaly}]{bouchekif2025islamic}
\bibinfo{author}{A.~Bouchekif}, \bibinfo{author}{S.~Rashwani},
  \bibinfo{author}{H.~Sbahi}, \bibinfo{author}{S.~Gaben},
  \bibinfo{author}{M.~Al-Khatib}, \bibinfo{author}{M.~Ghaly},
\newblock \bibinfo{title}{Assessing large language models on islamic legal
  reasoning: Evidence from inheritance law evaluation},
\newblock in: \bibinfo{booktitle}{Proceedings of The Second Arabic Natural
  Language Processing Conference (ArabicNLP 2025)},
  \bibinfo{publisher}{Association for Computational Linguistics},
  \bibinfo{address}{Suzhou, China}, \bibinfo{year}{2025}.
\bibitem[{Noureldien et~al.(2025)Noureldien, Suliman, Attallah, Mohamed, and
  Abdalla}]{noureldien-etal-2025-athar}
\bibinfo{author}{Y.~Noureldien}, \bibinfo{author}{H.~Suliman},
  \bibinfo{author}{F.~Attallah}, \bibinfo{author}{A.~Mohamed},
  \bibinfo{author}{S.~Abdalla},
\newblock \bibinfo{title}{Athar at {QIAS}2025: {LLM}-based question answering
  systems for islamic inheritance and classical islamic knowledge},
\newblock in: \bibinfo{booktitle}{Proceedings of The Third Arabic Natural
  Language Processing Conference: Shared Tasks},
  \bibinfo{publisher}{Association for Computational Linguistics},
  \bibinfo{address}{Suzhou, China}, \bibinfo{year}{2025}, pp.
  \bibinfo{pages}{914--922}.
\bibitem[{Alowaidi(2025)}]{alowaidi-2025-sea}
\bibinfo{author}{S.~Alowaidi},
\newblock \bibinfo{title}{{SEA}-team at {QIAS} 2025: Enhancing {LLM}s for
  question answering in islamic texts},
\newblock in: \bibinfo{editor}{K.~Darwish}, \bibinfo{editor}{A.~Ali},
  \bibinfo{editor}{I.~Abu~Farha}, \bibinfo{editor}{S.~Touileb},
  \bibinfo{editor}{I.~Zitouni}, \bibinfo{editor}{A.~Abdelali},
  \bibinfo{editor}{S.~Al-Ghamdi}, \bibinfo{editor}{S.~Alkhereyf},
  \bibinfo{editor}{W.~Zaghouani}, \bibinfo{editor}{S.~Khalifa},
  \bibinfo{editor}{B.~AlKhamissi}, \bibinfo{editor}{R.~Almatham},
  \bibinfo{editor}{I.~Hamed}, \bibinfo{editor}{Z.~Alyafeai},
  \bibinfo{editor}{A.~Alowisheq}, \bibinfo{editor}{G.~Inoue},
  \bibinfo{editor}{K.~Mrini}, \bibinfo{editor}{W.~Alshammari} (Eds.),
  \bibinfo{booktitle}{Proceedings of The Third Arabic Natural Language
  Processing Conference: Shared Tasks}, \bibinfo{publisher}{Association for
  Computational Linguistics}, \bibinfo{address}{Suzhou, China},
  \bibinfo{year}{2025}, pp. \bibinfo{pages}{940--946}. \URLprefix
  \url{https://aclanthology.org/2025.arabicnlp-sharedtasks.130/}.
\bibitem[{Singh et~al.(2025)Singh, Fry, Perelman, Tart, Ganesh, El-Kishky,
  McLaughlin, Low, Ostrow, Ananthram et~al.}]{singh2025openai}
\bibinfo{author}{A.~Singh}, \bibinfo{author}{A.~Fry},
  \bibinfo{author}{A.~Perelman}, \bibinfo{author}{A.~Tart},
  \bibinfo{author}{A.~Ganesh}, \bibinfo{author}{A.~El-Kishky},
  \bibinfo{author}{A.~McLaughlin}, \bibinfo{author}{A.~Low},
  \bibinfo{author}{A.~Ostrow}, \bibinfo{author}{A.~Ananthram}, et~al.,
\newblock \bibinfo{title}{Openai gpt-5 system card},
\newblock \bibinfo{journal}{arXiv preprint arXiv:2601.03267}
  (\bibinfo{year}{2025}).
\bibitem[{Anil et~al.(2023)Anil, Borgeaud, Alayrac, Yu, Soricut, Schalkwyk,
  Dai, Hauth, Millican et~al.}]{team2023gemini}
\bibinfo{author}{R.~Anil}, \bibinfo{author}{S.~Borgeaud},
  \bibinfo{author}{J.-B. Alayrac}, \bibinfo{author}{J.~Yu},
  \bibinfo{author}{R.~Soricut}, \bibinfo{author}{J.~Schalkwyk},
  \bibinfo{author}{A.~M. Dai}, \bibinfo{author}{A.~Hauth},
  \bibinfo{author}{K.~Millican}, et~al.,
\newblock \bibinfo{title}{Gemini: a family of highly capable multimodal
  models},
\newblock \bibinfo{journal}{arXiv preprint arXiv:2312.11805}
  (\bibinfo{year}{2023}).
\bibitem[{{DeepSeek AI}(2024)}]{deepseek2024r1}
\bibinfo{author}{{DeepSeek AI}}, \bibinfo{title}{DeepSeek-R1: Incentivizing
  Reasoning Capability in Large Language Models}, \bibinfo{type}{Technical
  Report}, DeepSeek AI, \bibinfo{year}{2024}. \bibinfo{note}{Technical Report}.
\bibitem[{Team et~al.(2025)Team, Abbas, Ahmad, Alam, Altinisik, Asgari,
  Boshmaf, Boughorbel, Chawla, Chowdhury
  et~al.}]{fanarteam2025fanararabiccentricmultimodalgenerative}
\bibinfo{author}{F.~Team}, \bibinfo{author}{U.~Abbas}, \bibinfo{author}{M.~S.
  Ahmad}, \bibinfo{author}{F.~Alam}, \bibinfo{author}{E.~Altinisik},
  \bibinfo{author}{E.~Asgari}, \bibinfo{author}{Y.~Boshmaf},
  \bibinfo{author}{S.~Boughorbel}, \bibinfo{author}{S.~Chawla},
  \bibinfo{author}{S.~Chowdhury}, et~al.,
\newblock \bibinfo{title}{Fanar: An arabic-centric multimodal generative ai
  platform},
\newblock \bibinfo{journal}{arXiv preprint arXiv:2501.13944}
  (\bibinfo{year}{2025}).
\bibitem[{Abbas et~al.(2026)Abbas, Ouzzani, Eltabakh, Sinan, Bhatia, Mubarak,
  Hawasly, Hashim, Darwish, and Alam}]{abbas2026fanar}
\bibinfo{author}{U.~Abbas}, \bibinfo{author}{M.~Ouzzani},
  \bibinfo{author}{M.~Y. Eltabakh}, \bibinfo{author}{O.~Sinan},
  \bibinfo{author}{G.~Bhatia}, \bibinfo{author}{H.~Mubarak},
  \bibinfo{author}{M.~Hawasly}, \bibinfo{author}{M.~Q. Hashim},
  \bibinfo{author}{K.~Darwish}, \bibinfo{author}{F.~Alam},
\newblock \bibinfo{title}{Fanar-sadiq: A multi-agent architecture for grounded
  islamic qa},
\newblock \bibinfo{journal}{arXiv preprint arXiv:2603.08501}
  (\bibinfo{year}{2026}).
\bibitem[{Yang et~al.(2025)Yang, Li, Yang, Zhang, Hui, Zheng, Yu, Gao, Huang,
  Lv et~al.}]{yang2025qwen3}
\bibinfo{author}{A.~Yang}, \bibinfo{author}{A.~Li}, \bibinfo{author}{B.~Yang},
  \bibinfo{author}{B.~Zhang}, \bibinfo{author}{B.~Hui},
  \bibinfo{author}{B.~Zheng}, \bibinfo{author}{B.~Yu},
  \bibinfo{author}{C.~Gao}, \bibinfo{author}{C.~Huang},
  \bibinfo{author}{C.~Lv}, et~al.,
\newblock \bibinfo{title}{Qwen3 technical report},
\newblock \bibinfo{journal}{arXiv preprint arXiv:2505.09388}
  (\bibinfo{year}{2025}).
\bibitem[{Cobbe et~al.(2021)Cobbe, Kosaraju, Bavarian, Chen, Jun, Kaiser,
  Plappert, Tworek, Hilton, Nakano, Hesse, and Schulman}]{abs-2110-14168}
\bibinfo{author}{K.~Cobbe}, \bibinfo{author}{V.~Kosaraju},
  \bibinfo{author}{M.~Bavarian}, \bibinfo{author}{M.~Chen},
  \bibinfo{author}{H.~Jun}, \bibinfo{author}{L.~Kaiser},
  \bibinfo{author}{M.~Plappert}, \bibinfo{author}{J.~Tworek},
  \bibinfo{author}{J.~Hilton}, \bibinfo{author}{R.~Nakano},
  \bibinfo{author}{C.~Hesse}, \bibinfo{author}{J.~Schulman},
\newblock \bibinfo{title}{Training verifiers to solve math word problems},
\newblock \bibinfo{journal}{CoRR} \bibinfo{volume}{abs/2110.14168}
  (\bibinfo{year}{2021}). \URLprefix \url{https://arxiv.org/abs/2110.14168}.
  \href{http://arxiv.org/abs/2110.14168}{{\tt arXiv:2110.14168}}.
\bibitem[{Hendrycks et~al.(2021)Hendrycks, Burns, Kadavath, Arora, Basart,
  Tang, Song, and
  Steinhardt}]{hendrycks2021measuringmathematicalproblemsolving}
\bibinfo{author}{D.~Hendrycks}, \bibinfo{author}{C.~Burns},
  \bibinfo{author}{S.~Kadavath}, \bibinfo{author}{A.~Arora},
  \bibinfo{author}{S.~Basart}, \bibinfo{author}{E.~Tang},
  \bibinfo{author}{D.~Song}, \bibinfo{author}{J.~Steinhardt},
  \bibinfo{title}{Measuring mathematical problem solving with the math
  dataset}, \bibinfo{year}{2021}. \URLprefix
  \url{https://arxiv.org/abs/2103.03874}.
  \href{http://arxiv.org/abs/2103.03874}{{\tt arXiv:2103.03874}}.
\bibitem[{Wei et~al.(2022)Wei, Wang, Schuurmans, Bosma, Xia, Chi, Le, Zhou
  et~al.}]{wei2023chainofthoughtpromptingelicitsreasoning}
\bibinfo{author}{J.~Wei}, \bibinfo{author}{X.~Wang},
  \bibinfo{author}{D.~Schuurmans}, \bibinfo{author}{M.~Bosma},
  \bibinfo{author}{F.~Xia}, \bibinfo{author}{E.~Chi}, \bibinfo{author}{Q.~V.
  Le}, \bibinfo{author}{D.~Zhou}, et~al.,
\newblock \bibinfo{title}{Chain-of-thought prompting elicits reasoning in large
  language models},
\newblock \bibinfo{journal}{Advances in Neural Information Processing Systems}
  \bibinfo{volume}{35} (\bibinfo{year}{2022}) \bibinfo{pages}{24824--24837}.
\bibitem[{Woo et~al.(2025)Woo, Chaleshtori, Marasovi{\'c}, and
  Marino}]{woo2025briefme}
\bibinfo{author}{J.~Woo}, \bibinfo{author}{F.~H. Chaleshtori},
  \bibinfo{author}{A.~Marasovi{\'c}}, \bibinfo{author}{K.~Marino},
\newblock \bibinfo{title}{Briefme: A legal nlp benchmark for assisting with
  legal briefs},
\newblock \bibinfo{journal}{arXiv preprint arXiv:2506.06619}
  (\bibinfo{year}{2025}).
\bibitem[{AlDahoul and Zaki(2025)}]{aldahoul-zaki-2025-nyuad-qias}
\bibinfo{author}{N.~AlDahoul}, \bibinfo{author}{Y.~Zaki},
\newblock \bibinfo{title}{{NYUAD} at {QIAS} shared task: Benchmarking the legal
  reasoning of {LLM}s in {A}rabic islamic inheritance cases},
\newblock in: \bibinfo{editor}{K.~Darwish}, \bibinfo{editor}{A.~Ali},
  \bibinfo{editor}{I.~Abu~Farha}, \bibinfo{editor}{S.~Touileb},
  \bibinfo{editor}{I.~Zitouni}, \bibinfo{editor}{A.~Abdelali},
  \bibinfo{editor}{S.~Al-Ghamdi}, \bibinfo{editor}{S.~Alkhereyf},
  \bibinfo{editor}{W.~Zaghouani}, \bibinfo{editor}{S.~Khalifa},
  \bibinfo{editor}{B.~AlKhamissi}, \bibinfo{editor}{R.~Almatham},
  \bibinfo{editor}{I.~Hamed}, \bibinfo{editor}{Z.~Alyafeai},
  \bibinfo{editor}{A.~Alowisheq}, \bibinfo{editor}{G.~Inoue},
  \bibinfo{editor}{K.~Mrini}, \bibinfo{editor}{W.~Alshammari} (Eds.),
  \bibinfo{booktitle}{Proceedings of The Third Arabic Natural Language
  Processing Conference: Shared Tasks}, \bibinfo{publisher}{Association for
  Computational Linguistics}, \bibinfo{address}{Suzhou, China},
  \bibinfo{year}{2025}, pp. \bibinfo{pages}{861--866}. \URLprefix
  \url{https://aclanthology.org/2025.arabicnlp-sharedtasks.118/}.
  \DOIprefix\doi{10.18653/v1/2025.arabicnlp-sharedtasks.118}.
\bibitem[{R{'}baiti et~al.(2025)R{'}baiti, El~Hachimi, Hmamouche, and
  El~Fallah~Seghrouchni}]{rbaiti-etal-2025-morai}
\bibinfo{author}{J.~R{'}baiti}, \bibinfo{author}{C.~El~Hachimi},
  \bibinfo{author}{Y.~Hmamouche}, \bibinfo{author}{A.~El~Fallah~Seghrouchni},
\newblock \bibinfo{title}{{M}or{AI} at {QIAS} 2025: Collaborative {LLM} via
  voting and retrieval-augmented generation for solving complex inheritance
  problems},
\newblock in: \bibinfo{editor}{K.~Darwish}, \bibinfo{editor}{A.~Ali},
  \bibinfo{editor}{I.~Abu~Farha}, \bibinfo{editor}{S.~Touileb},
  \bibinfo{editor}{I.~Zitouni}, \bibinfo{editor}{A.~Abdelali},
  \bibinfo{editor}{S.~Al-Ghamdi}, \bibinfo{editor}{S.~Alkhereyf},
  \bibinfo{editor}{W.~Zaghouani}, \bibinfo{editor}{S.~Khalifa},
  \bibinfo{editor}{B.~AlKhamissi}, \bibinfo{editor}{R.~Almatham},
  \bibinfo{editor}{I.~Hamed}, \bibinfo{editor}{Z.~Alyafeai},
  \bibinfo{editor}{A.~Alowisheq}, \bibinfo{editor}{G.~Inoue},
  \bibinfo{editor}{K.~Mrini}, \bibinfo{editor}{W.~Alshammari} (Eds.),
  \bibinfo{booktitle}{Proceedings of The Third Arabic Natural Language
  Processing Conference: Shared Tasks}, \bibinfo{publisher}{Association for
  Computational Linguistics}, \bibinfo{address}{Suzhou, China},
  \bibinfo{year}{2025}, pp. \bibinfo{pages}{947--952}.
  \DOIprefix\doi{10.18653/v1/2025.arabicnlp-sharedtasks.131}.
\bibitem[{Farouk~Zaki(2025)}]{farouk-zaki-2025-cis}
\bibinfo{author}{O.~Farouk~Zaki},
\newblock \bibinfo{title}{{CIS}-{RG} at {QIAS} 2025 shared task: Approaches for
  enhancing performance of {LLM} on islamic legal reasoning and its
  mathematical calculations},
\newblock in: \bibinfo{editor}{K.~Darwish}, \bibinfo{editor}{A.~Ali},
  \bibinfo{editor}{I.~Abu~Farha}, \bibinfo{editor}{S.~Touileb},
  \bibinfo{editor}{I.~Zitouni}, \bibinfo{editor}{A.~Abdelali},
  \bibinfo{editor}{S.~Al-Ghamdi}, \bibinfo{editor}{S.~Alkhereyf},
  \bibinfo{editor}{W.~Zaghouani}, \bibinfo{editor}{S.~Khalifa},
  \bibinfo{editor}{B.~AlKhamissi}, \bibinfo{editor}{R.~Almatham},
  \bibinfo{editor}{I.~Hamed}, \bibinfo{editor}{Z.~Alyafeai},
  \bibinfo{editor}{A.~Alowisheq}, \bibinfo{editor}{G.~Inoue},
  \bibinfo{editor}{K.~Mrini}, \bibinfo{editor}{W.~Alshammari} (Eds.),
  \bibinfo{booktitle}{Proceedings of The Third Arabic Natural Language
  Processing Conference: Shared Tasks}, \bibinfo{publisher}{Association for
  Computational Linguistics}, \bibinfo{address}{Suzhou, China},
  \bibinfo{year}{2025}, pp. \bibinfo{pages}{935--939}.
  \DOIprefix\doi{10.18653/v1/2025.arabicnlp-sharedtasks.129}.
\bibitem[{Bouchekif et~al.(2025)Bouchekif, Rashwani, Mohamed, Alkhatib, Sbahi,
  Gaben, Zaghouani, Erbad, and Ghaly}]{bouchekif-etal-2025-qias}
\bibinfo{author}{A.~Bouchekif}, \bibinfo{author}{S.~Rashwani},
  \bibinfo{author}{E.~S.~A. Mohamed}, \bibinfo{author}{M.~Alkhatib},
  \bibinfo{author}{H.~Sbahi}, \bibinfo{author}{S.~Gaben},
  \bibinfo{author}{W.~Zaghouani}, \bibinfo{author}{A.~Erbad},
  \bibinfo{author}{M.~Ghaly},
\newblock \bibinfo{title}{{QIAS} 2025: Overview of the shared task on islamic
  inheritance reasoning and knowledge assessment},
\newblock in: \bibinfo{booktitle}{Proceedings of The Third Arabic Natural
  Language Processing Conference: Shared Tasks},
  \bibinfo{publisher}{Association for Computational Linguistics},
  \bibinfo{address}{Suzhou, China}, \bibinfo{year}{2025}, pp.
  \bibinfo{pages}{851--860}.
  \DOIprefix\doi{10.18653/v1/2025.arabicnlp-sharedtasks.117}.
\bibitem[{Almasoud et~al.(2026)Almasoud, Al-Ghamdi, Alqifari, Alfear, and
  Al-Khalifa}]{ALMASOUD2026112589}
\bibinfo{author}{A.~Almasoud}, \bibinfo{author}{S.~Al-Ghamdi},
  \bibinfo{author}{R.~Alqifari}, \bibinfo{author}{N.~Alfear},
  \bibinfo{author}{H.~Al-Khalifa},
\newblock \bibinfo{title}{Mirathqa: A dataset for evaluating large language
  models on hanbali islamic inheritance reasoning tasks},
\newblock \bibinfo{journal}{Data in Brief}  (\bibinfo{year}{2026})
  \bibinfo{pages}{112589}.
  \DOIprefix\doi{https://doi.org/10.1016/j.dib.2026.112589}.
\bibitem[{AL-Smadi(2025)}]{al-smadi-2025-qu}
\bibinfo{author}{M.~AL-Smadi},
\newblock \bibinfo{title}{{QU}-{NLP} at {QIAS} 2025 shared task: A two-phase
  {LLM} fine-tuning and retrieval-augmented generation approach for islamic
  inheritance reasoning},
\newblock in: \bibinfo{editor}{K.~Darwish}, \bibinfo{editor}{A.~Ali},
  \bibinfo{editor}{I.~Abu~Farha}, \bibinfo{editor}{S.~Touileb},
  \bibinfo{editor}{I.~Zitouni}, \bibinfo{editor}{A.~Abdelali},
  \bibinfo{editor}{S.~Al-Ghamdi}, \bibinfo{editor}{S.~Alkhereyf},
  \bibinfo{editor}{W.~Zaghouani}, \bibinfo{editor}{S.~Khalifa},
  \bibinfo{editor}{B.~AlKhamissi}, \bibinfo{editor}{R.~Almatham},
  \bibinfo{editor}{I.~Hamed}, \bibinfo{editor}{Z.~Alyafeai},
  \bibinfo{editor}{A.~Alowisheq}, \bibinfo{editor}{G.~Inoue},
  \bibinfo{editor}{K.~Mrini}, \bibinfo{editor}{W.~Alshammari} (Eds.),
  \bibinfo{booktitle}{Proceedings of The Third Arabic Natural Language
  Processing Conference: Shared Tasks}, \bibinfo{publisher}{Association for
  Computational Linguistics}, \bibinfo{address}{Suzhou, China},
  \bibinfo{year}{2025}, pp. \bibinfo{pages}{892--898}.
  \DOIprefix\doi{10.18653/v1/2025.arabicnlp-sharedtasks.123}.
\bibitem[{Bekhouche et~al.(2025)Bekhouche, Sellam, Hichem, Distante, and
  Hadid}]{bekhouche2025cvpd}
\bibinfo{author}{S.~E. Bekhouche}, \bibinfo{author}{A.~Z. Sellam},
  \bibinfo{author}{T.~Hichem}, \bibinfo{author}{C.~Distante},
  \bibinfo{author}{A.~Hadid},
\newblock \bibinfo{title}{Cvpd at qias 2025 shared task: An efficient
  encoder-based approach for islamic inheritance reasoning},
\newblock in: \bibinfo{booktitle}{Proceedings of The Third Arabic Natural
  Language Processing Conference: Shared Tasks}, \bibinfo{year}{2025}, pp.
  \bibinfo{pages}{929--934}.
\bibitem[{Motasim~Hamed et~al.(2025)Motasim~Hamed, Ghneim, and
  Sonbol}]{motasim-hamed-etal-2025-hiast}
\bibinfo{author}{M.~Motasim~Hamed}, \bibinfo{author}{N.~Ghneim},
  \bibinfo{author}{R.~Sonbol},
\newblock \bibinfo{title}{{HIAST} at {QIAS} 2025: Retrieval-augmented {LLM}s
  with top-hit web evidence for {A}rabic islamic reasoning {QA}},
\newblock in: \bibinfo{booktitle}{Proceedings of The Third Arabic Natural
  Language Processing Conference: Shared Tasks},
  \bibinfo{publisher}{Association for Computational Linguistics},
  \bibinfo{address}{Suzhou, China}, \bibinfo{year}{2025}, pp.
  \bibinfo{pages}{883--891}.
  \DOIprefix\doi{10.18653/v1/2025.arabicnlp-sharedtasks.122}.
\bibitem[{Hossain and Afli(2025)}]{hossain-afli-2025-adapt-mtu}
\bibinfo{author}{S.~Hossain}, \bibinfo{author}{H.~Afli},
\newblock \bibinfo{title}{{ADAPT}{--}{MTU} {HAI} at {QIAS}2025: Dual-expert
  {LLM} fine-tuning and constrained decoding for {A}rabic islamic inheritance
  reasoning},
\newblock in: \bibinfo{editor}{K.~Darwish}, \bibinfo{editor}{A.~Ali},
  \bibinfo{editor}{I.~Abu~Farha}, \bibinfo{editor}{S.~Touileb},
  \bibinfo{editor}{I.~Zitouni}, \bibinfo{editor}{A.~Abdelali},
  \bibinfo{editor}{S.~Al-Ghamdi}, \bibinfo{editor}{S.~Alkhereyf},
  \bibinfo{editor}{W.~Zaghouani}, \bibinfo{editor}{S.~Khalifa},
  \bibinfo{editor}{B.~AlKhamissi}, \bibinfo{editor}{R.~Almatham},
  \bibinfo{editor}{I.~Hamed}, \bibinfo{editor}{Z.~Alyafeai},
  \bibinfo{editor}{A.~Alowisheq}, \bibinfo{editor}{G.~Inoue},
  \bibinfo{editor}{K.~Mrini}, \bibinfo{editor}{W.~Alshammari} (Eds.),
  \bibinfo{booktitle}{Proceedings of The Third Arabic Natural Language
  Processing Conference: Shared Tasks}, \bibinfo{publisher}{Association for
  Computational Linguistics}, \bibinfo{address}{Suzhou, China},
  \bibinfo{year}{2025}, pp. \bibinfo{pages}{923--928}.
  \DOIprefix\doi{10.18653/v1/2025.arabicnlp-sharedtasks.127}.
\bibitem[{Elrefai et~al.(2025)Elrefai, Lotfy~Elrefai, and
  Hassan~Esmail}]{elrefai-etal-2025-gumball}
\bibinfo{author}{E.~Elrefai}, \bibinfo{author}{M.~Lotfy~Elrefai},
  \bibinfo{author}{A.~Hassan~Esmail},
\newblock \bibinfo{title}{Gumball at {QIAS} 2025: {A}rabic {LLM} automated
  reasoning in islamic inheritance},
\newblock in: \bibinfo{booktitle}{Proceedings of The Third Arabic Natural
  Language Processing Conference: Shared Tasks},
  \bibinfo{publisher}{Association for Computational Linguistics},
  \bibinfo{address}{Suzhou, China}, \bibinfo{year}{2025}, pp.
  \bibinfo{pages}{953--959}. \URLprefix
  \url{https://aclanthology.org/2025.arabicnlp-sharedtasks.132/}.
  \DOIprefix\doi{10.18653/v1/2025.arabicnlp-sharedtasks.132}.

\end{thebibliography}

\appendix
\newpage
\newcolumntype{Y}{>{\RaggedRight\arraybackslash}X}
\newcolumntype{L}[1]{>{\RaggedRight\arraybackslash}p{#1}}

\newcolumntype{P}[1]{>{\RaggedRight\arraybackslash}p{#1}}

\section{Appendix: Heir Categories and
Kinship Coverage
}
\label{appendix/appendix_heirs}

\begin{figure*}[!t]
\centering
\includegraphics[width=0.9\textwidth]{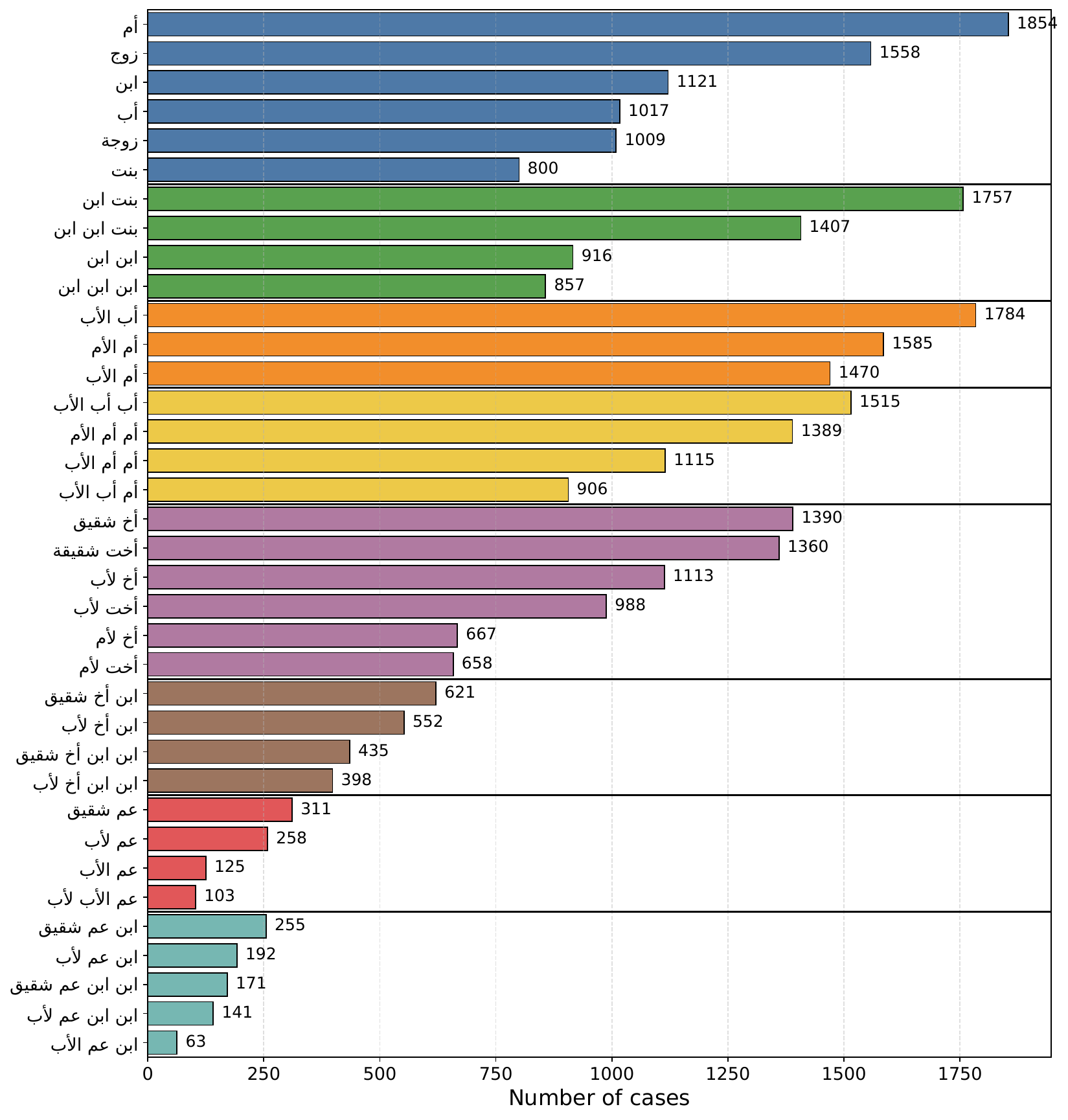}
\caption{Frequency distribution of all heir types in the corpus, ordered by kinship proximity.}
\label{fig:heir_distribution}
\end{figure*}

Islamic inheritance law defines a rich and structured set of potential heirs, reflecting varying degrees of kinship with the deceased. In our dataset, we explicitly cover all $36$ legally recognized heir categories as defined in classical Islamic inheritance law. These categories span a wide spectrum of kinship relations, ranging from close family members to more distant relatives.
\\
The dataset includes close relatives heirs who are directly related to the deceased, such as parents, spouses, and children. These close relatives typically appear with high frequency across inheritance cases and constitute the core of most inheritance configurations. They include ascendants (\textit{e.g.}, father, mother, and grandparents), descendants (e.g., sons, daughters, and grandchildren), as well as spouses. In addition to these direct relatives, collateral relatives are also considered, including siblings and half-siblings (full, paternal, or maternal). Under Islamic inheritance law, these heirs become legally relevant in many scenarios, particularly when direct ascendants or descendants are absent, and their entitlement is governed by well-defined blocking and prioritization rules. For analysis, we group the $37$ heir categories into $8$ \emph{kinship levels}, ordered by proximity to the deceased (Level~1: closest relatives; Level~8: most distant extended-family relatives); see Figure~\ref{fig:heir_distribution}. As shown in the figure, heirs from Levels~1--3 are the most frequent in the corpus, whereas distant heirs from higher levels occur much less often (Levels~6--8).
\\
More generally, Islamic inheritance law recognizes that all legally defined heirs may become entitled to inherit depending on the absence of closer relatives. Accordingly, extended relatives across multiple generational levels—such as nephews, uncles, cousins, and their descendants—play a legally significant role in inheritance cases where nearer heirs are absent. In such situations, inheritance is determined by hierarchical succession rules, making these cases particularly suitable for evaluating long-range dependency reasoning and complex blocking mechanisms.

\begin{table*}[!t]
\centering
\small
\setlength{\tabcolsep}{8pt}
\renewcommand{\arraystretch}{1.10}

\begin{tabular}{P{0.12\textwidth} P{0.26\textwidth} P{0.18\textwidth} P{0.30\textwidth}}
\toprule
\textbf{Arabic heir} & \textbf{English gloss} & \textbf{Arabic heir} & \textbf{English gloss} \\
\midrule

\arcell{أم} & Mother & \arcell{أخ شقيق} & Full brother \\
\arcell{زوج} & Husband & \arcell{أخت شقيقة} & Full sister \\
\arcell{ابن} & Son & \arcell{أخ لأب} & Paternal half-brother (same father) \\
\arcell{أب} & Father & \arcell{أخت لأب} & Paternal half-sister (same father) \\
\arcell{زوجة} & Wife & \arcell{أخ لأم} & Maternal half-brother (same mother) \\
\arcell{بنت} & Daughter & \arcell{أخت لأم} & Maternal half-sister (same mother) \\
\addlinespace[3pt]

\arcell{بنت ابن} & Son's daughter (granddaughter) & \arcell{ابن أخ شقيق} & Son of a full brother (full nephew) \\
\arcell{بنت ابن ابن} & Son's son's daughter (great-granddaughter) & \arcell{ابن أخ لأب} & Son of a paternal half-brother (paternal nephew) \\
\arcell{ابن ابن} & Son's son (grandson) & \arcell{ابن ابن أخ شقيق} & Son of a full nephew (grand-nephew) \\
\arcell{ابن ابن ابن} & Son's son's son (great-grandson) & \arcell{ابن ابن أخ لأب} & Son of a paternal nephew (grand-nephew) \\
\addlinespace[3pt]

\arcell{أب الأب} & Paternal grandfather (father's father) & \arcell{عم شقيق} & Full paternal uncle (father's full brother) \\
\arcell{أم الأم} & Maternal grandmother (mother's mother) & \arcell{عم لأب} & Paternal half-uncle (father's half-brother) \\
\arcell{أم الأب} & Paternal grandmother (father's mother) & \arcell{عم الأب} & Father's full paternal uncle (paternal grand-uncle) \\
\arcell{أب أب الأب} & Paternal great-grandfather (father's father's father) & \arcell{عم الأب لأب} & Father's paternal half-uncle (paternal grand-uncle) \\
\arcell{أم أم الأم} & Maternal great-grandmother (mother's mother's mother) & \arcell{ابن عم شقيق} & Son of a full paternal uncle (paternal cousin) \\
\arcell{أم أم الأب} & Paternal great-grandmother (father's mother's mother) & \arcell{ابن عم لأب} & Son of a paternal half-uncle (paternal cousin) \\
\arcell{أم أب الأب} & Paternal great-grandmother (father's father's mother) & \arcell{ابن ابن عم شقيق} & Son of a paternal cousin (cousin's son) \\
 & & \arcell{ابن ابن عم لأب} & Son of a paternal cousin (cousin's son) \\
 & & \arcell{ابن عم الأب} & Son of father's full paternal uncle (father's cousin) \\

\bottomrule
\end{tabular}
\caption{Arabic--English glosses for the $36$ heir categories used in our corpus.}
\label{tab:heirs_gloss_4col}
\end{table*}

\clearpage
\section{Illustrative Example (Arabic)}
\label{app:arabic_example}
The following example, taken from our corpus, provides a detailed step-by-step illustration of Islamic inheritance calculation. It explains the reasoning process from the identification of the heirs, the determination of the prescribed shares and the ʿaṣabah, and the taʾṣīl and correction of the case, to the final verification that the total of the distributed shares equals the full estate. For clarity and faithfulness to the source material, we provide both the original Arabic version and its English translation.
\begin{tcolorbox}[myarbox,breakable]
\begin{otherlanguage*}{arabic}
\noindent\textbf{Question:}\par
\medskip
مات وترك: أخوين شقيقين، وخمس أخوات شقيقات، وأب الأب، وأربع بنات ابن ابن.\par
ما نصيب كل وارث؟
\end{otherlanguage*}
\end{tcolorbox}

\begin{tcolorbox}[myarbox,breakable]
\begin{otherlanguage*}{arabic}
\noindent\textbf{Answer:}\par

\noindent\textLR{\texttt{<think>}}

\medskip
في هذه المسألة، المتوفى ترك اثنين إخوة أشقاء، وخمس أخوات شقيقات، وأب الأب، وأربع بنات ابن ابن.  
لنتبع خطوات الحل المعتمد لتحديد نصيب كل وارث.

\medskip
\textbf{الخطوة الأولى: تحديد نصيب كل وارث من أصحاب الفروض وتعيين العصبات}

نبدأ بتحديد نصيب كل وارث من التركة بحسب الفروض الشرعية، ومن ثم يُعطى الباقي للعصبة إن وجدوا:

\medskip
\textbf{بنات ابن الابن الأربع:}

نصيبهن هو الثلثان (2/3) من التركة، وذلك لأن بنات الابن ـ ومثلهن بنات ابن الابن ـ يُعاملن معاملة البنات عند عدم وجود من يحجبهن، كالبنت الصلبية أو ابن الصلب أو ابن ابن أعلى منهن. وترث الواحدة النصف، وإذا كنّ أكثر من واحدة فلهن الثلثان، بشرط عدم وجود ابن ابن في درجتهن يعصبهن.

وقد قال الله تعالى: «يُوصِيكُمُ اللَّهُ فِي أَوْلَادِكُمْ لِلذَّكَرِ مِثْلُ حَظِّ الأُنثَيَيْنِ فَإِنْ كُنَّ نِسَاءً فَوْقَ اثْنَتَيْنِ فَلَهُنَّ ثُلُثَا مَا تَرَكَ وَإِنْ كَانَتْ وَاحِدَةً فَلَهَا النِّصْفُ».

\medskip

\textbf{أب الأب --الجد الصحيح:}

نصيبه هو السدس \textLR{(1/6)} من التركة، لأنه وُجد مع إخوة ومع وجود صاحب فرض، فيرث بأفضل الأحوال الثلاثة: المشاركة، أو ثلث الباقي، أو سدس المال كله. وفي هذه المسألة كان سدس المال كله هو الأوفر له، فيرث السدس.

\medskip
\textbf{الإخوة الأشقاء الاثنان والأخوات الشقيقات الخمس:}

يرثون ما بقي من التركة تعصيبًا، ويُقسم بينهم للذكر مثل حظ الأنثيين، لأنهم عصبة بالنفس ذكورًا وإناثًا.

\medskip
\textbf{الخطوة الثانية: تأصيل المسألة}

بنات ابن الابن لهن الثلثان \textLR{(2/3)}.  
أب الأب له السدس \textLR{(1/6)}.  

مقامات الفروض هي \textLR{3} و\textLR{6}، فيكون أصل المسألة هو \textLR{6}.

توزيع السهام على أصل المسألة:
\begin{itemize}[noitemsep, topsep=0pt, partopsep=0pt]
\item بنات ابن الابن الأربع: \textLR{4} أسهم.
\item أب الأب: \textLR{1} سهم.
\item مجموع سهام أصحاب الفروض = \textLR{5} أسهم.
\item الباقي = \textLR{6 - 5 = 1} سهم، وهو نصيب العصبة.
\end{itemize}

\medskip
\textbf{الخطوة الثالثة: تصحيح المسألة وإخراج نصيب كل فرد}

سهم العصبة \textLR{(1)} يحتاج إلى التقسيم على اثنين إخوة أشقاء وخمس أخوات شقيقات.

نحسب عدد الوحدات وفق قاعدة: للذكر مثل حظ الأنثيين:

\textLR{(2} إخوة \textLR{× 2) + (5} أخوات \textLR{× 1) = 9}

وبما أن السهم لا ينقسم على \textLR{9}، نُصحح المسألة بضرب أصلها في عدد الوحدات:

جامعة المسألة = \textLR{6 × 9 = 54} سهمًا.

\medskip
\textbf{توزيع السهام بعد التصحيح:}

\textbf{1- بنات ابن الابن الأربع:}

\textLR{4 × 9 = 36} سهمًا، أي \textLR{9} أسهم لكل بنت.  
النسبة المئوية: \textLR{(9/54) × 100\% = 16.67\%} لكل بنت.

\medskip
\textbf{2- أب الأب:}

\textLR{1 × 9 = 9} أسهم.  
النسبة المئوية: \textLR{(9/54) × 100\% = 16.67\%}.

\medskip
\textbf{3- العصبة (الإخوة والأخوات الأشقاء):}

لهم \textLR{9} أسهم على \textLR{9} وحدات:
\begin{itemize}[noitemsep, topsep=0pt, partopsep=0pt]
\item كل أخ شقيق: \textLR{2} سهمين.
\item النسبة المئوية: \textLR{(2/54) × 100\% = 3.7\%}.
\item كل أخت شقيقة: \textLR{1} سهم.
\item النسبة المئوية: \textLR{(1/54) × 100\% = 1.85\%}.
\end{itemize}

\medskip
\textbf{خلاصة الأنصبة الفردية:}
\begin{itemize}[noitemsep, topsep=0pt, partopsep=0pt]
\item كل بنت ابن ابن: \textLR{9} أسهم \textLR{(16.67\%)}.
\item أب الأب: \textLR{9} أسهم \textLR{(16.67\%)}.
\item كل أخ شقيق: \textLR{2} سهم \textLR{(3.7\%)}.
\item كل أخت شقيقة: \textLR{1} سهم \textLR{(1.85\%)}.
\end{itemize}

وبجمع الأنصبة:

\textLR{(4 × 9) + (1 × 9) + (2 × 2) + (5 × 1)}  
\textLR{= 36 + 9 + 4 + 5 = 54} سهمًا،

وبذلك يستقيم التقسيم دون نقص أو زيادة.

\medskip
\noindent\textLR{\texttt{</think>}}

\end{otherlanguage*}
\end{tcolorbox}

\begin{tcolorbox}[myarbox,breakable]
\begin{otherlanguage*}{arabic}

\noindent\textLR{\texttt{<answer>}}

\medskip
\textbf{الجواب:}

بناءً على المعطيات الواردة في المسألة، والتي تفيد بوفاة المورّث وترك: أربع بنات ابن ابن، وأب الأب، واثنين إخوة أشقاء، وخمس أخوات شقيقات؛ فإن التركة تُقسَّم على النحو الآتي وفقًا لأحكام الشريعة الإسلامية:

\medskip
\textbf{1. بنات الابن الابن الأربع:}

يستحقن الثلثين من كامل التركة، كما جاء في قوله تعالى: «فَإِنْ كُنَّ نِسَاءً فَوْقَ اثْنَتَيْنِ فَلَهُنَّ ثُلُثَا مَا تَرَكَ». وعليه، يكون نصيب كل واحدة من بنات الابن الابن تسعة أجزاء من أربعة وخمسين جزءًا \textLR{(9/54)} من التركة.

\medskip
\textbf{2. أب الأب (الجد):}

يستحق سدس التركة فرضًا. وعليه، يكون نصيبه تسعة أجزاء من أربعة وخمسين جزءًا \textLR{(9/54)} من التركة.

\medskip
\textbf{3. الإخوة الأشقاء والأخوات الشقيقات:}

يشتركون في باقي التركة تعصيبًا، للذكر منهم مثل حظ الأنثيين. وعليه:
\begin{itemize}[noitemsep, topsep=0pt, partopsep=0pt]
\item \textbf{كل أخ من الأخوين الشقيقين:} يستحق جزأين من أربعة وخمسين جزءًا \textLR{(2/54)} من التركة.
\item \textbf{كل أخت من الأخوات الشقيقات الخمس:} تستحق جزءًا واحدًا من أربعة وخمسين جزءًا \textLR{(1/54)} من التركة.
\end{itemize}

\medskip
\textbf{وعليه، يكون نصيب كل وارث بالتفصيل كما يأتي:}
\begin{itemize}[noitemsep, topsep=0pt, partopsep=0pt]
\item \textbf{لكل بنت من بنات الابن الابن الأربع:} \textLR{(9/54)} من التركة.
\item \textbf{لأب الأب:} \textLR{(9/54)} من التركة.
\item \textbf{لكل أخ من الأخوين الشقيقين:} \textLR{(2/54)} من التركة.
\item \textbf{لكل أخت من الأخوات الشقيقات الخمس:} \textLR{(1/54)} من التركة.
\end{itemize}

\medskip
\noindent\textLR{\texttt{</answer>}}

\end{otherlanguage*}
\end{tcolorbox}

\textbf{Illustrative Example (English)}
\label{app:english_example}

\begin{tcolorbox}[myarbox,breakable]
\textbf{Question:}

A person died leaving: two full brothers, five full sisters, the paternal grandfather (the deceased’s father’s father; \textit{al-jadd al-ṣaḥīḥ}), and four granddaughters through the son’s line (daughters of the grandson). What is the share of each heir?
\end{tcolorbox}

\begin{tcolorbox}[myarbox,breakable]

\noindent\texttt{<think>}

\medskip
In this case, the deceased left two full brothers, five full sisters, the paternal grandfather, and four granddaughters through the son’s line. Let us follow the adopted steps in order to determine the share of each heir.

\medskip
\textbf{Step One: Assign the fixed shares (\textit{aṣḥāb al-furūḍ}) and identify the residuaries (\textit{ʿaṣabah})}

We begin by assigning the prescribed fixed shares to those entitled to them. Any remainder of the estate is then given to the \textit{ʿaṣabah} (agnatic residuaries), if present.

\medskip
\textbf{1) Four granddaughters through the son’s line (daughters of the grandson):}

Their collective share is two-thirds \((2/3)\) of the estate. Granddaughters in the male line are treated like daughters when there is no closer descendant who blocks them---such as a direct son, a direct daughter in certain configurations, or a nearer male-line grandson (a son’s son of a higher degree). A single such granddaughter would inherit one-half; two or more inherit two-thirds, provided there is no male-line grandson at the same degree who would render them residuaries with him (\textit{taʿṣīb bi-l-ghayr}).

Allah the Exalted says (Q 4:11): ``Allah instructs you concerning your children: for the male is the share of two females. If there are two or more females, then for them is two-thirds of what he left; and if there is one, then for her is one-half.''

\medskip
\textbf{2) The paternal grandfather (\textit{al-jadd al-ṣaḥīḥ}):}

His assessed entitlement is one-sixth \((1/6)\) of the estate. When the paternal grandfather is present alongside full siblings and there are fixed-share heirs in the case, he is assessed under the rule that he receives whichever is greatest of: (i) one-sixth of the whole estate, (ii) one-third of the remainder after the fixed shares, or (iii) sharing the residuary remainder with the siblings (\textit{muqāsamah}). In this case, one-sixth of the whole estate is the most advantageous, so he takes one-sixth.

\medskip
\textbf{3) Two full brothers and five full sisters:}

They take the remainder of the estate as residuaries (\textit{taʿṣīb}), distributed with each male receiving the portion of two females.

\medskip
\textbf{Step Two: Establish the root of the case (\textit{taʾṣīl al-masʾalah})}

The granddaughters through the son’s line are entitled to \(2/3\), and the paternal grandfather is entitled to \(1/6\). The denominators are \(3\) and \(6\), so the root of the case is \(6\).

Distribution of shares on the root \((6)\):

\begin{itemize}[noitemsep, topsep=0pt, partopsep=0pt]
\item Granddaughters through the son’s line: \(4\) shares (i.e., \(2/3\) of \(6\)).
\item Paternal grandfather: \(1\) share (i.e., \(1/6\) of \(6\)).
\item Total shares of the fixed-share heirs (\textit{aṣḥāb al-furūḍ}): \(5\) shares.
\item Remainder: \(6 - 5 = 1\) share (for the residuaries).
\end{itemize}

\medskip
\textbf{Step Three: Correct the case (\textit{taṣḥīḥ}) and derive each individual’s share}

The residuary remainder \((1\) share on a root of \(6)\) must be divided among two full brothers and five full sisters. We compute the entitlement units according to the rule ``the male receives the share of two females'': \((2 \text{ brothers} \times 2) + (5 \text{ sisters} \times 1) = 9\) units. Since \(1\) share cannot be divided evenly into \(9\) units, we correct the case by multiplying the root by \(9\): \(6 \times 9 = 54\) total shares.

\medskip
\textbf{Distribution after correction (root = \(54\)):}

\textbf{1) Four granddaughters through the son’s line:}

They receive \(4 \times 9 = 36\) shares, i.e., \(9\) shares for each granddaughter.\\
Percentage for each: \((9/54) \times 100\% = 16.67\%\).

\medskip
\textbf{2) Paternal grandfather:}

He receives \(1 \times 9 = 9\) shares.\\
Percentage: \((9/54) \times 100\% = 16.67\%\).

\medskip
\textbf{3) Residuaries (two full brothers and five full sisters):}

They collectively receive \(9\) shares to be distributed across \(9\) entitlement units:
\begin{itemize}[noitemsep, topsep=0pt, partopsep=0pt]
\item Each full brother: \(2\) shares \((2/54 = 3.70\%)\).
\item Each full sister: \(1\) share \((1/54 = 1.85\%)\).
\end{itemize}

\medskip
\textbf{Summary of the individual shares:}
\begin{itemize}[noitemsep, topsep=0pt, partopsep=0pt]
\item Each granddaughter through the son’s line: \(9/54\) of the estate \((16.67\%)\).
\item Paternal grandfather: \(9/54\) of the estate \((16.67\%)\).
\item Each full brother: \(2/54\) of the estate \((3.70\%)\).
\item Each full sister: \(1/54\) of the estate \((1.85\%)\).
\end{itemize}

\textbf{Verification:}

By summing the distributed shares: \((4 \times 9) + (1 \times 9) + (2 \times 2) + (5 \times 1) = 36 + 9 + 4 + 5 = 54\) shares. Thus, the division is sound, without deficiency or excess.

\medskip
\noindent\texttt{</think>}

\end{tcolorbox}

\begin{tcolorbox}[myarbox,breakable]

\noindent\texttt{<answer>}

\medskip
\textbf{Answer:}

Based on the stated heirs---four granddaughters through the son’s line (daughters of the grandson), the paternal grandfather (\textit{al-jadd al-ṣaḥīḥ}), two full brothers, and five full sisters---the estate is divided as follows according to the rules applied in this example.

\medskip
\textbf{1) Four granddaughters through the son’s line:}

They are collectively entitled to two-thirds of the estate. Each granddaughter receives \(9/54\) of the estate.

\medskip
\textbf{2) Paternal grandfather:}

He is entitled to one-sixth of the estate by prescribed share. He receives \(9/54\) of the estate.

\medskip
\textbf{3) Two full brothers and five full sisters:}

They take the remainder by \textit{taʿṣīb}, with each male receiving the share of two females.
\begin{itemize}[noitemsep, topsep=0pt, partopsep=0pt]
\item \textbf{Each full brother:} receives \(2/54\) of the estate.
\item \textbf{Each full sister:} receives \(1/54\) of the estate.
\end{itemize}

\medskip
\textbf{Accordingly, the share of each heir in detail is as follows:}
\begin{itemize}[noitemsep, topsep=0pt, partopsep=0pt]
\item \textbf{Each of the four granddaughters through the son’s line (daughters of the grandson):} \(9/54\) of the estate.
\item \textbf{The paternal grandfather (the deceased’s father’s father):} \(9/54\) of the estate.
\item \textbf{Each of the two full brothers:} \(2/54\) of the estate.
\item \textbf{Each of the five full sisters:} \(1/54\) of the estate.
\end{itemize}

\medskip
\noindent\texttt{</answer>}

\end{tcolorbox}
\section{Reasoning Prompt (English Translation)}
\label{app:reasoning_prompt}

The following prompt is the English translation of the original Arabic reasoning prompt used in our experiments. It specifies the step-by-step reasoning process that the model is expected to follow in order to solve Islamic inheritance problems according to the majority school (\textit{madhhab al-jumhūr}), from identifying the heirs and determining those who are blocked, to assigning the prescribed shares, identifying \textit{taʿṣīb}, determining whether the case involves \textit{ʿawl} or \textit{radd}, applying correction when needed, and providing the final distribution of the estate with the percentage share of each heir.

\begin{tcolorbox}[myarbox,breakable]

You are a jurist specialized in \textit{ʿilm al-farāʾiḍ} according to \textit{madhhab al-jumhūr}, and your task is to solve the case with precise Sharʿī accuracy.

This list contains the heirs extracted from the text of the case (it is not allowed to add, remove, or modify any heir or their number):\\
\texttt{\{heirs\}}

\medskip
\textbf{The case:}\\
\texttt{\{question\}}

\medskip
\textbf{What is required in the answer:}\\
Formulate the case briefly, and clarify who inherits and who is blocked, mentioning the reason for the blocking if any. If possible, include a Sharʿī proof for inheritance from the Qurʾān or the Sunnah. Then explain the ruling of each heir: whether he or she is a holder of a prescribed share (\textit{ṣāḥib farḍ}) or an heir by \textit{taʿṣīb}, while stating the fraction in the form \texttt{a/b} if it is a prescribed share. Also clarify whether the case involves \textit{ʿawl} or \textit{radd}, mention the correction (\textit{taṣḥīḥ}) if any, and then conclude the answer by stating the final distribution of the estate together with the percentage share of each heir.

\medskip
\textbf{Mandatory rule:}\\
If the expression \foreignlanguage{arabic}{باقي التركة} (the remainder of the estate) or \foreignlanguage{arabic}{كل التركة} (the whole estate) is mentioned, and the entitled party consists of more than one heir, then the way the estate is divided among them must be explained, and the share of each heir must be stated explicitly. It is not permissible to suffice with a general statement.

\medskip
\textbf{Important note:}\\
Write the solution in one paragraph or in several connected paragraphs, using a clear scholarly juristic style.

\end{tcolorbox}
\section{Structured Solution Format}
\label{app:answer_structured_format}
To support deterministic and fine grained evaluation, each model output includes, in addition to
the free-form \texttt{<think>} and the concise \texttt{<answer>}, a structured field
\texttt{answer\_structured}. This field provides machine-readable intermediate results that mirror the
standard inheritance workflow used by legal experts and can be parsed automatically.

\medskip
\noindent
\textbf{General schema.} \\
For each instance, \texttt{answer\_structured} is a JSON object with the following components:

\begin{itemize}[itemsep=0pt, topsep=0pt, parsep=0pt, partopsep=0pt]
  \item \textbf{\texttt{heirs}}: the list of heirs who actually inherit (eligible heirs), each with
  \texttt{heir} (Arabic label) and \texttt{count} (multiplicity).
  \item \textbf{\texttt{blocked}}: the list of present but excluded heirs (blocked by \textit{hajb}),
  each with \texttt{heir} and \texttt{count}.
  \item \textbf{\texttt{shares}}: the initial allocation \emph{before} any global adjustment
  (\textit{ʿawl} or \textit{radd}). Each entry specifies \texttt{heir}, \texttt{count}, and a
  \texttt{fraction} string.
  \item \textbf{\texttt{awl\_or\_radd}}: a categorical flag indicating whether a global adjustment is applied:
\texttt{"none"} / \texttt{"awl"} / \texttt{"radd"} (\foreignlanguage{arabic}{لا} / \foreignlanguage{arabic}{عول} / \foreignlanguage{arabic}{رد}).
  \item \textbf{\texttt{awl\_stage}}: represents the state of the case after adjustment. It is included
only when \texttt{awl\_or\_radd} is \texttt{"awl"} or \texttt{"radd"}.
  \item \textbf{\texttt{post\_tasil}}: the final normalized distribution (\textit{taṣīl}). It contains
  final denominator (\texttt{total\_shares}) and per-head final results via  \texttt{per\_head\_shares} and \texttt{per\_head\_percent}.
\end{itemize}

\medskip
\noindent
\textbf{Conventions.}
Heir names must follow the benchmark taxonomy (Arabic labels). Counts must be explicit whenever an heir
type appears multiple times. Fractions are represented as strings (e.g., \texttt{"1/6"}, \texttt{"2/3"},
\texttt{"3/24"}). When no adjustment is needed, \texttt{awl\_or\_radd} is \texttt{"none"},
\texttt{asl\_after\_awl} is \texttt{null}.
\\
This structured representation makes inheritance reasoning measurable at multiple levels: eligibility
and blocking decisions (\texttt{heirs}/\texttt{blocked}), initial share assignment (\texttt{shares}),
correct handling of global adjustment (\texttt{awl\_stage}), and the final
normalized distribution (\texttt{post\_tasil}).

\medskip
\noindent
\textbf{Template (case-general JSON).}
\\
\Needspace{18\baselineskip}
\begin{otherlanguage*}{english}
\begin{jsonbox}
"answer_structured": {
  "heirs": [
    {
      "heir": "heir1", "count": INT
    },
    {
      "heir": "heir2", "count": INT
    }
  ],
  "blocked": [
    {
      "heir": "blocked1", "count": INT
    },
    {
      "heir": "blocked2", "count": INT
    }
  ],
  "shares": [
    {
       "heir": "heir1", "count": INT,  "fraction": "a/b"
    },
    {
       "heir": "heir2", "count": INT, "fraction": "a/b"
    }
  ],
  "awl_or_radd": "none" | "awl" | "radd",
  "awl_stage": {
    "asl_after_awl": INT | null,
    "distribution": [
      {
        "heir": "heir1", 
        "count": INT, 
        "fraction": "a/b"
      }
    ]
  },
  "post_tasil": {
    "total_shares": INT,
    "distribution": [
      {
        "heir": "heir1",
        "count": INT,
        "per_head_shares": "a/b",
        "per_head_percent": FLOAT
      }
    ]
  }
}
\end{jsonbox}
\end{otherlanguage*}
\par
\medskip
\noindent

\end{document}